\documentclass{article}

\usepackage{microtype}
\usepackage{graphicx}
\usepackage{multirow}
\usepackage{graphicx}
\usepackage{subfigure}
\usepackage{booktabs} 
\usepackage[table,xcdraw]{xcolor}
\usepackage{tcolorbox}
\definecolor{cGreen}{HTML}{2e75b5}
\definecolor{cgray}{HTML}{FAFAFA}
\definecolor{blue}{HTML}{0055cc}
\definecolor{red}{HTML}{cc1100}
\definecolor{orange}{HTML}{cc7700}
\definecolor{green}{HTML}{339955}
\definecolor{Highlight}{rgb}{0.12,0.49,0.85}
\definecolor{my_red}{HTML}{ff0000}

\usepackage{hyperref}

\usepackage[accepted]{icml2025}

\usepackage{amsmath}
\usepackage{amssymb}
\usepackage{mathtools}
\usepackage{amsthm}
\usepackage{url}
\usepackage{caption}
\usepackage{wrapfig}
\usepackage[capitalize,noabbrev]{cleveref}
\usepackage{pifont}
\setcitestyle{circle}
\crefname{section}{Sec.}{Secs.}
\Crefname{section}{Section}{Sections}
\Crefname{table}{Table}{Tables}
\crefname{table}{Tab.}{Tabs.}
\crefname{equation}{Eq.}{Eqs.}
\Crefname{equation}{Eq.}{Eqs.}
\crefname{figure}{Fig.}{Figs.}
\Crefname{figure}{Fig.}{Figs.}
\crefname{algorithm}{Algo.}{Algo.}
\Crefname{algorithm}{Algo.}{Algos.}

\usepackage{xspace}

\newcommand{\INT}{\texttt{INT4}\xspace}
\newcommand{\FP}{\texttt{FP16}\xspace}
\newcommand{\NAME}{IntLoRA\xspace}
\newcommand{\MUL}{IntLoRA$\rm{_{MUL}}$\xspace}
\newcommand{\SHIFT}{IntLoRA$\rm{_{SHIFT}}$\xspace}

\theoremstyle{plain}

\theoremstyle{definition}

\theoremstyle{remark}

\DeclareMathOperator{\sign}{sign}
\usepackage[textsize=tiny]{todonotes}

\icmltitlerunning{IntLoRA: Integral Low-rank Adaptation of Quantized Diffusion Models}

\begin{document}

\twocolumn[
\icmltitle{IntLoRA: Integral Low-rank Adaptation of Quantized Diffusion Models}



\icmlsetsymbol{lead}{*}
\icmlsetsymbol{correspondence}{$\dagger$}

\begin{icmlauthorlist}
\icmlauthor{Hang Guo}{thu}
\icmlauthor{Yawei Li}{ethz,lead,correspondence}
\icmlauthor{Tao Dai}{szu,correspondence}
\icmlauthor{Shu-Tao Xia}{thu,pclab}
\icmlauthor{Luca Benini}{ethz}
\end{icmlauthorlist}

\icmlaffiliation{thu}{Tsinghua University}
\icmlaffiliation{ethz}{ETH Z\"{u}rich}
\icmlaffiliation{szu}{Shenzhen University}
\icmlaffiliation{pclab}{Peng Cheng Laboratory}

\icmlcorrespondingauthor{Yawei Li}{yawei.li@vision.ee.ethz.ch}
\icmlcorrespondingauthor{Tao Dai}{daitao.edu@gmail.com}

\icmlkeywords{Low-rank Adaptation, Network Quantization, Diffusion Models}

\vskip 0.3in
]



\printAffiliationsAndNotice{}  

\begin{abstract}
Fine-tuning pre-trained diffusion models under limited budgets has gained great success. In particular, the recent advances that directly fine-tune the quantized weights using Low-rank Adaptation (LoRA) further reduces training costs. Despite these progress, we point out that existing adaptation recipes are not inference-efficient. Specifically, additional post-training quantization (PTQ) on tuned weights is needed during deployment, which results in noticeable performance drop when the bit-width is low. Based on this observation, we introduce IntLoRA, which adapts quantized diffusion models with integer-type low-rank parameters, to include inference efficiency during tuning. Specifically, IntLoRA enables pre-trained weights to remain quantized during training, facilitating fine-tuning on consumer-level GPUs. During inference, IntLoRA weights can be seamlessly merged into pre-trained weights to directly obtain quantized downstream weights without PTQ. Extensive experiments show our IntLoRA achieves significant speedup on both training and inference without losing performance. Code is available at \href{https://github.com/csguoh/IntLoRA}{https://github.com/csguoh/IntLoRA}.
\end{abstract}

\section{Introduction}

Recently, large-scale text-to-image diffusion models~\citep{rombach2022high,saharia2022photorealistic,podell2023sdxl} have shown promising capabilities for image generation. Taking advantage of the strong generative prior of pre-trained parameters, a range of downstream adaptation applications have emerged, such as subject-driven generation~\citep{ruiz2023dreambooth}, style-customized generation~\citep{sohn2023styledrop}, and controllable generation~\citep{zhang2023adding}. However, fully fine-tuning large pre-trained models for downstream tasks poses challenges on consumer-level GPUs. For instance, only loading the $\texttt{FP32}$ FLUX.1-dev~\citep{flux} weights into GPUs can consume over 24GB of memory, let alone subsequent fine-tuning. Therefore, the huge fine-tuning costs hinder personalized diffusion model customization.

\begin{figure}[!t]
    \centering
    \includegraphics[width=0.96\linewidth]{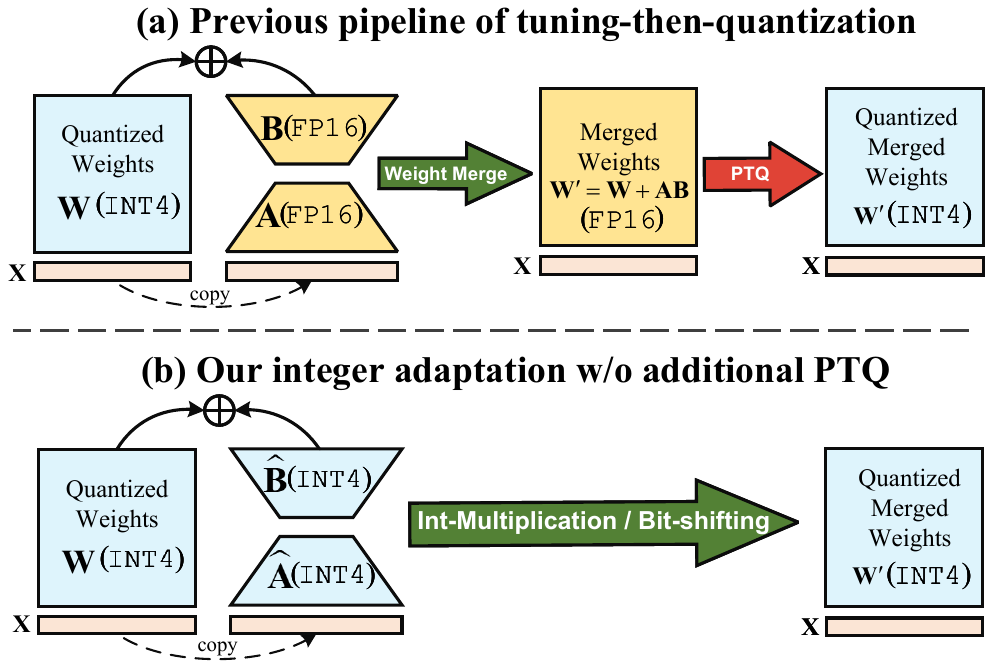}
    \vspace{-2mm}
    \caption{(a) The arithmetic inconsistency between the pre-trained and adaptation weights leads to the merged weights still in \FP. Consequently, additional PTQ is needed for low-bit inference. (b) Our IntLoRA allows to work directly on \INT arithmetic, ensuring the merged weights seamlessly in \INT format and streamlining the whole process.}
    \vspace{-3mm}
    \label{fig:motivation}
\end{figure}

To facilitate efficient training, recent advances have introduced parameter efficient fine-tuning (PEFT)~\citep{houlsby2019parameter,jia2022visual} techniques, such as LoRA~\citep{hu2021lora}, to fine-tune a limited number of parameters. With the reduced gradient and optimizer states, they can achieve comparable or even better adaptation performance than fully fine-tuning. More recently, some works~\citep{dettmers2024qlora,qin2024ir-qlora} 
have successfully married PEFT and network quantization to allow the low-rank adaptation directly on the quantized weights (as shown in~\cref{fig:motivation}(a)). Through reducing the bit-widths, the GPU costs during fine-tuning are further decreased.

\begin{figure}[!t]
    \centering
    \includegraphics[width=0.98\linewidth]{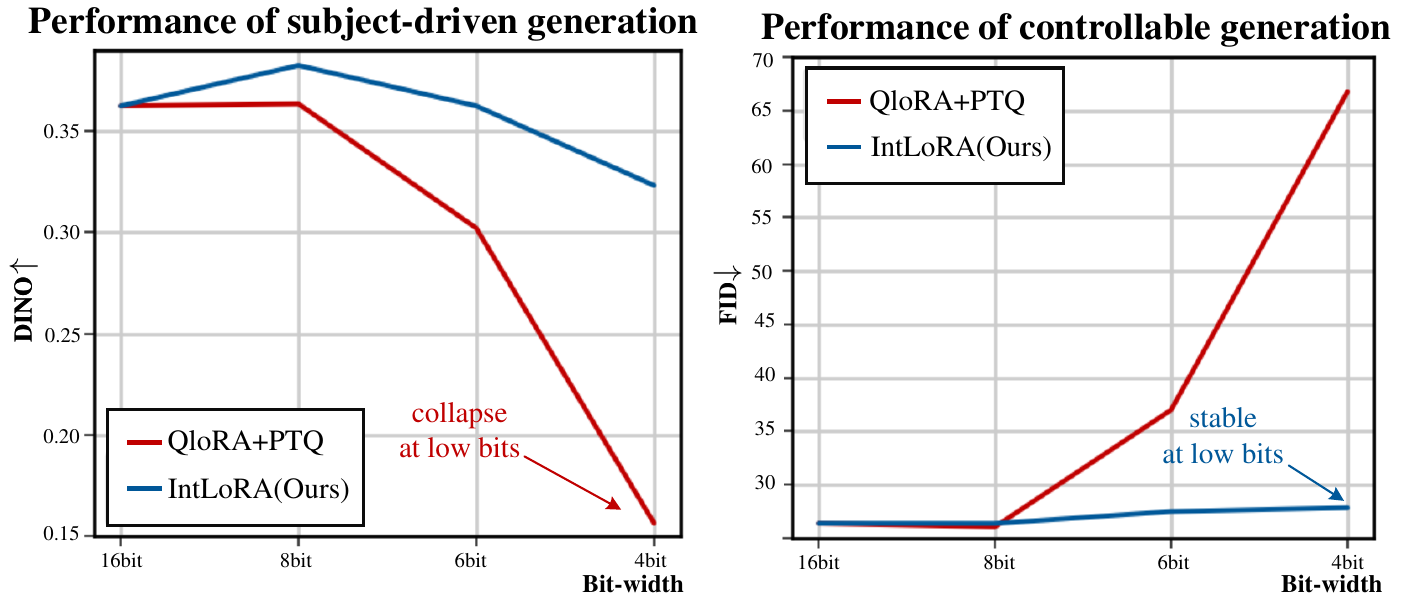}
    \vspace{-2mm}
    \caption{The utilization of PTQ on the downstream merged weights leads to severe performance degradation under low bit-width quantization.}
    \vspace{-6mm}
    \label{fig:motivation2}
\end{figure}

Although the reduced training costs have facilitated user customization, obtaining an inference-aware tuning recipe remains an open challenge. Specifically, existing methods predominantly employ floating-point (\textit{e.g.}, \FP) low-rank parameters during training, as a result, it is inevitable to convert the quantized pre-trained weights back to \FP for arithmetic consistency to merge low-rank weights into pre-trained weights. During test-time, this pipeline necessitates additional post-training quantization (PTQ) on the \FP merged weights for accelerated inference, which is pipeline-complicated and incurs significant performance drop when the bit-width is low (see \cref{fig:motivation2}).

To address these challenges, a potential solution is to also transfer the adaptation weights to integer arithmetic. In this way, all weights during fine-tuning are in integers, thus ensuring the merged weights naturally being quantized. Despite these promising properties, it is non-trivial to accurately quantize the low-rank weights. For example, while zero initializing low-rank weights are advantageous for fine-tuning~\citep{hu2021lora}, it poses quantization challenges due to substantial quantization errors from small values. Furthermore, the additive form of the original LoRA forces the pre-trained and adaptation weights to share the same quantizer for seamless weight merging, which restricts available parameter space during fine-tuning.

In this work, we propose \NAME, which achieves integral low-rank parameters for \textit{both training and inference efficient} diffusion models. In detail, we introduce the Adaptation-Quantization Separation (AQS) technique, which employs a task-agnostic auxiliary matrix to enable quantization-friendly low-rank parameters without disrupting the gradient trajectory of the original LoRA. Additionally, we present the Multiplicative Low-rank Adaptation (MLA), which reformulates the mathematical structure of LoRA from addition to multiplication. This remains mathematically equivalent to the original but allows for independent optimization of adaptation weights. Furthermore, we develop the Variance Matching Control (VMC) to align the pre-trained and auxiliary matrices. For implementation, we provide two versions, \textit{i.e.}, \MUL, and \SHIFT. The \MUL learns quantized low-rank parameters and can be seamlessly merged through integer multiplication, while \SHIFT introduces log2-quantization and operates by bit-shifting the quantized weights for downstream adaptation. We evaluate our \NAME on various diffusion personalization tasks. Extensive experiments show that \NAME presents  impressive efficiency and performance.


\section{Related Work}

\noindent
\textbf{Parameter-efficient fine-tuning of diffusion models.}
In order to reduce the fine-tuning cost of large models, parameter-efficient fine-tuning (PEFT) has recently gained great interests~\citep{lian2022scaling,chavan2023one,li2021prefix,he2021towards,jie2023fact}. For example, prompt-based methods~\citep{jia2022visual} append learnable prompts to modify the input space. Adapter-based methods~\citep{houlsby2019parameter,chen2022adaptformer} employ additional bottleneck structures as bypass branches for adaptation. Moreover, LoRA~\citep{hu2021lora} adopts low-rank matrices to learn the weight updates for downstream tasks. In this work, we mainly focus on LoRA since it has been widely applied in diffusion model fine-tuning and can be merged into pre-trained weights without increasing inference costs.

\noindent
\textbf{Network quantization of diffusion models.}
Quantization~\citep{nagel2021white} is an effective technique to speed deep-learning models and can be categorized into quantization-aware training (QAT)~\citep{jacob2018quantization,li2024qdm,li2022qvit,xu2023q-detr} and post-training quantization (PTQ)~\citep{wang2023towards,nahshan2021loss,li2021brecq,wei2022qdrop,liu2023pd,huang2024slim}. In the context of diffusion model quantization, existing works mainly focus on PTQ because of the significant overhead of retraining diffusion models. For example, PTQ4DM~\citep{shang2023ptq4dm} makes the first attempt to quantize diffusion models to 8 bits. After that, Q-Diffusion~\citep{li2023qdiff} further achieves improved performance and lower bit-width. EfficientDM~\citep{he2023efficientdm} introduces LoRA to fine-tune the pre-trained model to allow comparable performance with QAT. TFMQ-DM~\citep{huang2024tfmq} proposes to quantize the time-embedding layer individually for better performance.

\noindent
\textbf{Joint PEFT and quantization for efficient fine-tuning.}
Benefiting from the scaling law, the pre-trained models have become increasingly large, which makes even loading models challenging. To allow fine-tuning on consumer-level GPUs for user customization, some work attempts to apply PEFT techniques directly on the quantized pre-trained weights. Specifically, QLoRA~\citep{dettmers2024qlora} proposes to quantize the LLMs before fine-tuning the LLMs with LoRA. Despite the reduced GPU usage during training due to the import of only the quantized model, QLoRA does not maintain quantized at inference since the quantized weights need to be converted to \FP again so as to be merged with the LoRA weights. QA-LoRA~\citep{xu2023qalora} develops a group-wise quantization through sharing parameters across channels but at the cost of impairing the adaptation ability. IR-QLoRA~\citep{qin2024ir-qlora} analyzes the entropy loss of quantization from an information theory view, but it also needs to convert the quantized weights back to \FP during inference.

\section{Preliminary}
The LoRA~\citep{hu2021lora} introduces a low-rank matrix $\Delta \mathbf{W}$ to learn the weight increments for adapting the pre-trained weights $\mathbf{W}\in \mathbb{R}^{C_{out}\times C_{in}}$ to downstream tasks. In implementation, the $\Delta \mathbf{W}$ is formulated as the matrix multiplication of two low-rank matrices $\mathbf{A} \in \mathbb{R}^{C_{out} \times d} $ and $\mathbf{B} \in  \mathbb{R}^{d \times C_{in}}$, where the inner dimension $d$ is the pre-defined rank. During fine-tuning, the pre-trained weight $\mathbf{W}$ is frozen and only the low-rank $\mathbf{A}, \mathbf{B}$ are trainable. Since $d\ll \min\{C_{in}, C_{out}\}$, the number of trainable parameters can be very small compared to full fine-tuning, thus reducing the GPU footprint of gradients and optimizer states. The output during downstream fine-tuning is calculated as $\mathbf{y} = \mathbf{W}\mathbf{x} +  \lambda \cdot (\mathbf{A}\mathbf{B})\mathbf{x}$, where $\lambda$ is the LoRA scale to adjust the control strength. During inference, the task-specific $\mathbf{AB}$ can naturally be  merged into the pre-trained weights, \textit{i.e.}, $\mathbf{W'} = \mathbf{W} + \lambda \cdot \mathbf{A}\mathbf{B}$, without increasing additional costs.

Even though the LoRA can alleviate training costs through reduced gradients and optimizer states, it still needs to load huge \FP pre-trained weights. Given the increasing pre-trained model size, it becomes impractical to only use LoRA to fine-tune the diffusion models on consumer-level GPUs. To further reduce the training memory, recent advancements~\citep{dettmers2024qlora,xu2023qalora,qin2024ir-qlora} have introduced network quantization to allow direct fine-tuning on the integer weights. Formally, given a tensor $\mathbf{X}$, the target bit-width $b$, the quantization process can be defined as: 
\begin{equation}    
\label{eq:quant-folrmular}
\Hat{\mathbf{X}} = s \cdot (\text{clip}(\lfloor\frac{\mathbf{X}}{s}\rceil+z,0,2^b -1)-z) \triangleq s\cdot(\mathbf{X}_\mathrm{round}-z),
\end{equation}
\noindent
where $\lfloor \cdot \rceil$ is the $\mathrm{round}$ function, $s=\frac{\max(\mathbf{X})-\min(\mathbf{X})}{2^b-1}$ is the scaling factor, and $z=-\lfloor \frac{\min(\mathbf{X})}{s} \rceil$ is the zero-point.

Despite current methods allow user to train customized models under a low memory budget, they all require additional PTQ on the fine-tuned weights for fast inference, which leads to noticeable performance degradation when the quantization bit-width is low.

\begin{figure*}[!t]
    \centering
    \includegraphics[width=0.95\linewidth]{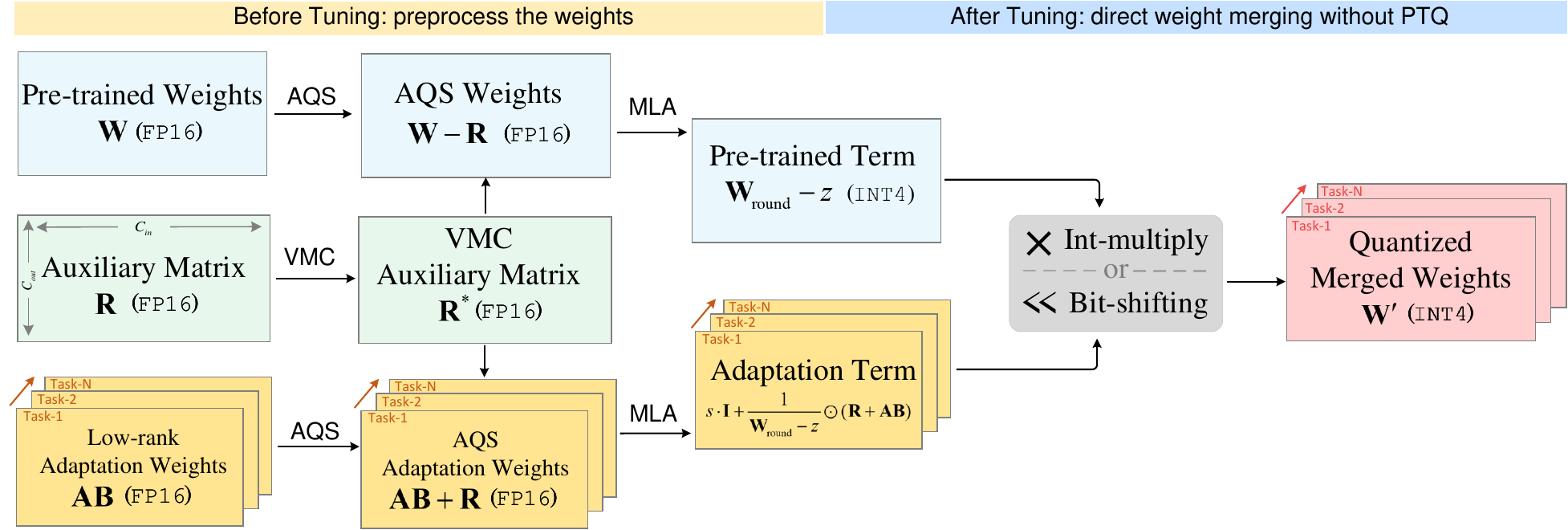}
    \vspace{-1mm}
    \caption{ {\textbf{Before tuning}, we propose the Adaptation Quantization Separation (AQS) to incorporate auxiliary matrix into pre-trained weights and low-rank weights for zero-initialized but quantization-friendly distribution. Then, the Multiplicative Low-rank Adaptation (MLA) is used to reformulate additive LoRA into the product of the ``pre-training term'' and the ``adaptation term''. At last, we introduce the Variance Matching Control (VMC) to adjust the distribution of the adaptation term by modulating the auxiliary matrix. \textbf{After tuning}, we use hardware-friendly integer multiplication or bit shifting to directly generate quantized merged weights without additional PTQ.} The detailed algorithm is given in~\cref{sec:suppl-algo}.}
    \vspace{-4mm}
    \label{fig:pipeline}
\end{figure*}

\section{Methodology}

In this work, we aim to remove the additional PTQ of the merged weights by introducing integer-type low-rank parameters during fine-tuning. In this way, both the $\mathbf{AB}$ and $\mathbf{W}$ are in the same arithmetic type, thus ensuring the merged $\mathbf{W}'$ is naturally already quantized. However, several technical challenges arise when transferring LoRA to integer arithmetic. First, the $\mathbf{AB}$ in the original LoRA is zero-initialized to ensure the behavior of the model is similar to the pre-trained one at the beginning of training. Although helpful for fine-tuning, this initialization complicates the quantization process. For instance, the all-zero distribution requires a separately designed quantizer at the beginning of tuning,  {since the scaling factor $s = 0$ leads to an infinite $\frac{\mathbf{X}}{s}$ in~\cref{eq:quant-folrmular}}. Second, the vanilla LoRA merges the \FP $\mathbf{AB}$ and $\mathbf{W}$ using addition. When both $\mathbf{AB}$ and $\mathbf{W}$ are quantized, it is essential to ensure that they share identical quantization parameters to enable PTQ-free weight merging. This requirement leads to constrained parameter space, thus limiting the adaptation ability.

\subsection{Integral Low-rank Adaptation}
\label{sec:method}
To address the above challenges, we propose \NAME to operate adaptation on the integer arithmetic. The overall pipeline is shown in \cref{fig:pipeline}.

\textbf{Adaptation-quantization separation.}
The vanilla LoRA adopts zero initialization on the adaptation parameter $\mathbf{AB}$. Although this strategy can improve performance, the all-zero distribution is not quantization-friendly. To allow accurate quantization while maintaining the correct gradient, we propose the Adaptation-Quantization-Separation (AQS) mechanism. The key observation is that 
\textit{the adaptation requires gradients from zero-initialized weights while the quantization does not}. Therefore, we can split the adaptation weights into the gradient-enabled zero part and the gradient-free nonzero part. Formally, let $\mathbf{R}$ be the auxiliary matrix to serve as the nonzero part, $\mathcal{Q}$  be the quant-dequant operator, then our AQS can be formulated as:
\begin{equation}
\label{eq:quant-peft-separate}
    \mathbf{W'} = \mathcal{Q}[\mathbf{W} - \mathrm{sg}(\mathbf{R})] + \mathrm{sg}(\mathbf{R}) + \mathbf{AB},
\end{equation}
\noindent
where $\mathrm{sg}(\cdot)$ denotes the stop gradient operation. Thanks to the AQS, the $\mathbf{AB}$ can be zero-initialized for the same gradient as the original LoRA, while $\mathrm{sg}({\mathbf{R}}) + \mathbf{AB}$ facilitate subsequent quantization  {by specifically designing the auxiliary matrix $\mathbf{R}$ as discussed in~\cref{sec:ablation}}. In the following part, we will ignore the $\mathrm{sg}(\cdot)$ notation for clarity.

\textbf{Multiplicative low-rank adaptation.}
The vanilla LoRA employs additive form $\mathbf{W} + \mathbf{AB}$ for weight merge. However, it is difficult to seamlessly fuse the quantized $\hat{\mathbf{W}}$ and $\hat{\mathbf{AB}}$ when they are quantized by independent quantizers. To this end, we propose Multiplicative Low-rank Adaptation (MLA) to rewrite the form of the original LoRA into a quantization-friendly multiplication form. Specifically, denote the quant-dequant results as $\mathcal{Q}(\mathbf{W} - \mathbf{R}) = s\cdot (\mathbf{W}_\mathrm{round}-z)$, then the MLA can be derived as follows:
\begin{equation}
\label{eq:mul-lora}
\begin{aligned}
\mathbf{W'} &= \mathcal{Q}(\mathbf{W}- \mathbf{R})+ \mathbf{R}+\mathbf{AB} \\ 
   &= s\cdot(\mathbf{W}_\mathrm{round} -z) +  \mathbf{R} + \mathbf{AB}\\
   &= \textcolor[RGB]{192, 0, 0}{[s\cdot \mathbf{I}+\frac{1}{\mathbf{W}_\mathrm{round}-z}\odot(\mathbf{ R+AB})]} \odot \textcolor[RGB]{0,0,192}{(\mathbf{W}_\mathrm{round}-z)},
\end{aligned}
\end{equation}
\noindent
where the task-specific \textcolor[RGB]{192, 0, 0}{adaptation term} is trainable and will be quantized, and the \textcolor[RGB]{0,0,192}{pre-trained term} is already in integer type and is shared across tasks. $\mathbf{I}$ is an all-one matrix. The operator $\odot$ denotes the Hadamard product of two matrices. The proposed MLA is mathematically equivalent to its additive counterpart, while is more quantization-friendly since it avoids the shared quantizer of pre-trained and adaptation weights. It is noteworthy that the adaptation term is still in \FP at this step, and we will detail its quantization strategies in \cref{sec:implementation}.

\textbf{Variance matching control.}
One opportunity brought from the multiplicative form in \cref{eq:mul-lora} is that we can apply the log2-quantization on the adaptation term, thus allowing more efficient bit-shifting on the pre-trained term. However, log2-quantization is notoriously more difficult than common uniform quantization~\citep{nagel2021white} and requires appropriate distribution properties, \textit{e.g.}, most values concentrated around zero to allow for the utilization of as many quantization bins as possible on the logarithmic scale. Here, we revisit the adaptation term in \cref{eq:mul-lora} aiming to find useful mathematical insights. Given the $\mathbf{AB}$ is orders of magnitude smaller than $\mathbf{R}$  {(the justification is shown in~\cref{sec:suppl-ab-less-than-R})}, we approximate the adaptation term in \cref{eq:mul-lora} by removing $\mathbf{AB}$ from it, namely,
\begin{equation}
\begin{aligned}
s\cdot \mathbf{I} +\frac{\mathbf{R}}{\mathbf{W}_\mathrm{round}-z} &= s \cdot \mathbf{I} + \frac{s\cdot \mathbf{R}}{s \cdot (\mathbf{W}_\mathrm{round}-z)} \\
&\approx s\cdot \mathbf{I} + \frac{s\cdot \mathbf{R}}{\mathbf{W-R}} 
= \frac{s\cdot \mathbf{W}}{\mathbf{W}-\mathbf{R}}.
\end{aligned}
\end{equation}
From this derivation, it follows that the auxiliary matrix $\mathbf{R}$ is crucial for controlling the distribution shape of the adaptation term. Unfortunately, we find there exists a dilemma in choosing an appropriate distribution for $\mathbf{R}$. On one hand, it is desirable for the values in $\mathbf{R}$ to be larger. Formally, let $\sigma_{\mathbf{R}}$ be the variance of the element in $\mathbf{R}$ which is a random variable, it can be derived the expectation of the adaptation term converges to zero when $\sigma_{\mathbf{R}}$ approaches  infinity, namely,
\begin{equation}
\begin{aligned}
    \mathbb{E} \left[\lim_{\sigma_\mathbf{R}\to\infty} s\cdot \mathbf{I} + \frac{\mathbf{R}}{\mathbf{W-R}} \right] 
    =  \mathbb{E} \left[ \lim_{\sigma_\mathbf{R}\to\infty} \frac{s\cdot \mathbf{W}}{\mathbf{W}-\mathbf{R}}\right] = 0. \label{eqn:zero_mean}
\end{aligned}
\end{equation}
On the other hand, setting $\sigma_{\mathbf{R}}$ too large can also lead the $\mathcal{Q}(\mathbf{W-R})$ uncorrelated to the original $\mathbf{W}$, \textit{i.e}, namely,
\begin{equation}
    \lim_{\sigma_\mathbf{R}\to\infty} \rho(\mathcal{Q}(\mathbf{W} - \mathbf{R}), \mathbf{W}) = \lim_{\sigma_R\to\infty} \frac{\sigma_\mathbf{W}}{\sqrt{\sigma_\mathbf{W}^2 + \sigma_\mathbf{R}^2}} = 0,
    \label{eq:low-correlation}
\end{equation}
where the $\rho(\cdot, \cdot)$ denotes the correlation coefficient. ~\cref{eq:low-correlation} indicates that a over-large $\sigma_\mathbf{R}$ makes it difficult to reconstruct the original signal $\mathbf{W}$ through dequantizing the $\mathcal{Q}(\mathbf{W-R})$ due to the low correlation. In short, it is important to choose an appropriate $\sigma_{\mathbf{R}}$ to strike a balance between quantization difficulty and information retention. We also give the visualization of this choice dilemma  in~\cref{fig:distribution_viz_ablation}. 
To this end, we propose the Variance Matching Control (VMC) mechanism. Specifically, we first multiply $\mathbf{R}$ by the variance ratio $r = \frac{\sigma_{\mathbf{W}}}{\sigma_{\mathbf{R}}} \in \mathbb{R}^{C_{out}}$ for rough alignment from $\mathbf{R}$ to the scale of $\mathbf{W}$. After that, we introduce a scalar $\alpha$ as an exponent of $r$, \textit{i.e.}, $r^{\alpha}$, to fine-grain the search for the optimal $\mathbf{R}^{*}$. As a result, the variance-matched auxiliary matrix can be denoted as $\mathbf{R}^{*} = r^{\alpha} \cdot \mathbf{R}$, and we can use this to obtain the distribution suitable for log2-quantization. Since $r^{\alpha}$ can be shared across tasks, it is only of negligible cost. In addition to the $\sigma_{\mathbf{R}}$, we observe the distribution shape of $\mathbf{R}$ also has an effect on performance, and we give a detailed discussion in \cref{sec:ablation}. It should be noted that the $\mathbf{R}$ can be online generated during fine-tuning using the distribution statistics and fixed random seed, thus avoiding the need to store its  \FP parameters.

\begin{table*}[!tb]
\centering
\caption{Quantitative comparison on subject-driven generation tasks. The notion ``WxAy" represents the bit-widths of weights ``W" and activations ``A". The best results are \textbf{bolded}.}
\label{tab:compare-db}
\vspace{-2mm}
\setlength{\tabcolsep}{14pt}
\scalebox{0.97}{
\begin{tabular}{@{}l|c|cccc@{}}
\toprule
methods       & nbits & DINO$\uparrow$   & CLIP-I$\uparrow$ & CLIP-T$\uparrow$ & LPIPS$\downarrow$  \\ \midrule
LoRA~\citep{hu2021lora}  & W16A16  & 0.4828 & 0.6968 & 0.2954 & 0.8076 \\
\midrule
QLoRA~\citep{dettmers2024qlora}   & W8A8       & 0.4153 & 0.6661 & 0.2824 & 0.8088 \\
QA-LoRA~\citep{xu2023qalora}       & W8A8       & 0.4156 & 0.6664 & 0.2834 & 0.8086 \\
IR-QLoRA~\citep{qin2024ir-qlora}      & W8A8       & 0.4070 & 0.6630 & 0.2841 & 0.8110 \\
\rowcolor[HTML]{EFEFEF} 
\SHIFT (Ours) & W8A8 & 0.4353 & 0.6842 & 0.2841 & 0.8257 \\
\rowcolor[HTML]{EFEFEF} 
\MUL (Ours)  & W8A8  & \textbf{0.4498} & \textbf{0.6882} & \textbf{0.2858} & \textbf{0.8062} \\
\midrule
QLoRA~\citep{dettmers2024qlora}       & W4A8       & 0.2136 & 0.6134 & 0.2510 & 0.8201 \\
QA-LoRA~\citep{xu2023qalora}       & W4A8       & 0.4127 & 0.6897 & 0.2700 & 0.8281 \\
IR-QLoRA~\citep{qin2024ir-qlora}      & W4A8       & 0.3722 & 0.6719 & 0.2707 & 0.8186 \\
\rowcolor[HTML]{EFEFEF} 
\SHIFT (Ours) & W4A8   & 0.4039 & 0.6716 & 0.2709 & 0.8187 \\ 
\rowcolor[HTML]{EFEFEF} 
\MUL (Ours)  & W4A8    & \textbf{0.4242} & \textbf{0.6913} & \textbf{0.2710} & \textbf{0.8181} \\
\bottomrule
\end{tabular}%
}
\end{table*}

\subsection{Implementation of IntLoRA}
\label{sec:implementation}
Benefiting from the quantization-friendly weight distribution, we can implement our IntLoRA with two variants according to different quantizers on the adaptation term. The first variant employs the uniform affine quantizer on the adaptation term, thus enabling weight merge through integer-type multiplication. The second variant introduces the more hardware-friendly log2 quantizer to achieve downstream adaptation by bit-shifting the quantized pre-trained weights. More details are given below.

\textbf{Integer multiplication form.}
We employ uniform affine quantization on the adaptation term, with the scaling factor and zero-point denoted as $\bar{s}$ and $\bar{z}$, and the quantized results as $\mathbf{U}_\mathrm{round}$, 
 then our \MUL can be formalized as:
\begin{equation}
\mathbf{W'} = \bar{s}\cdot(\mathbf{U}_\mathrm{round}-\bar{z})\odot(\mathbf{W}_\mathrm{round}-z).
\end{equation}
\noindent
\textbf{Bit-shifting form.}
Denote the adaptation term in \cref{eq:mul-lora} as $\mathbf{V}$ for clarity, we first compute the bit shift value as follows:
\begin{equation}
    \mathrm{shift} = \text{clip}(\lfloor-\log_2|\mathbf{V}|\rceil,0,2^b -1).
\end{equation}
Then the weight adaptation with \SHIFT can be represented as:
\begin{equation}
\begin{aligned}
\mathbf{W'} &= \sign(\mathbf{V})\odot 2^{-\mathrm{shift}} \odot(\mathbf{W}_\mathrm{round}-z)\\ &= \sign(\mathbf{V}) \odot [(\mathbf{W}_\mathrm{round}-z)\gg \mathrm{shift}] \\
&= \frac{1}{2^N} \odot \sign(\mathbf{V}) \odot [(\mathbf{W}_\mathrm{round}-z)\ll(N-\mathrm{shift})],
\end{aligned}
\end{equation}
\noindent
where $\sign(\mathbf{V}) \in \{-1,1\}$ and $N=2^b-1$. Since the direct right-shifting on $\mathbf{W}_\mathrm{round}-z$ may lead to truncation error, we thus use $N-\mathrm{shift}$ with a scaling factor $\frac{1}{2^N}$ to equivalently convert to the left-shifting for error reduction.

\section{Experiments}

\subsection{Experimental Setup}

\noindent
\textbf{Datasets.}
We evaluate on multiple adaptation tasks, including subject-driven generation~\citep{ruiz2023dreambooth}, controllable generation~\citep{zhang2023adding}, and style-customized image generation~\citep{sohn2023styledrop}. For the subject-driven generation, we use a subset which contains 15 text-subject pairs from the Dreambooth~\citep{ruiz2023dreambooth} dataset, for fast training and evaluation. For controllable generation, we consider three sub-tasks, \textit{i.e}, Segmentation map to Image (S2I) on ADE20K dataset~\citep{zhou2017ade}, Landmark to Face (L2F) on CelebA-HQ dataset~\citep{karras2017progressive}, and the Canny edge to Image (C2I) on the COCO dataset~\citep{lin2014microsoft}. For the style-customized generation, we employ the StyleDrop~\citep{sohn2023styledrop} dataset, which includes 18 style images, and we use 6 text prompts for each style to generate images with style similar to the style image and content aligned with the text prompt.

\noindent
\textbf{Implementation details.}
We employ the StableDiffusionV1.5~\citep{rombach2022high} as the pre-trained backbone for subject-driven generation and controllable generation. We further employ larger SDXL~\citep{podell2023sdxl} as the pre-trained model in the style-customized generation. We use uniform quantization~\citep{nagel2021white} to quantize the weights per-channel and activations per-tensor. Since previous methods mainly focus on efficient training, with the tuned weights still in \FP, to make a fair comparison, we apply additional PTQ on the merged weights for efficient inference. As for the training of the quantized adaptation term, we use the Straight Through Estimator (STE) to allow back-propagation. For the proposed variance matching control, we employ the ratio of the maxima of the sampled distributions as a fast estimator for the variance. Due to the page limit, we provide more details in~\cref{sec:suppl-additoinal-details}.

\begin{figure*}[!tb]
    \centering
    \vspace{-3mm}
    \includegraphics[width=0.92\linewidth]{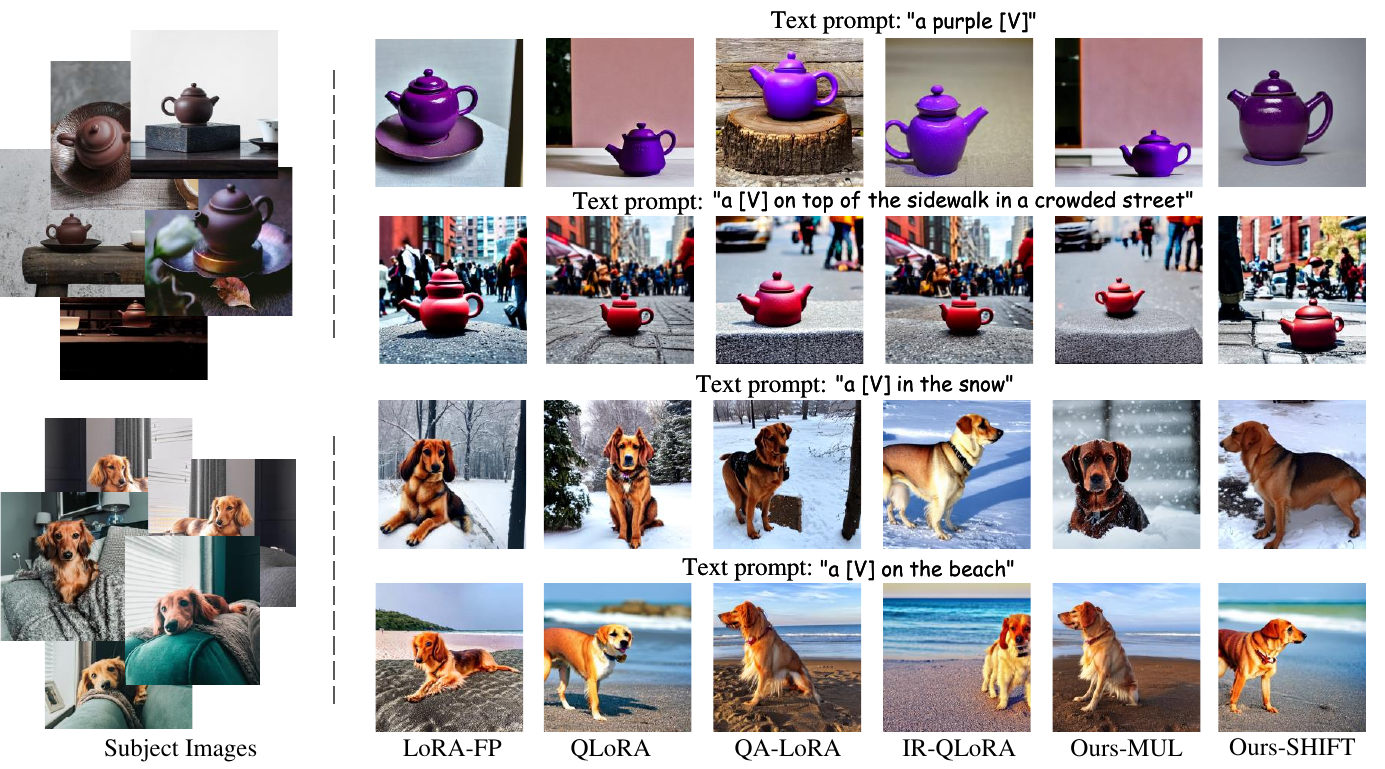} 
    \vspace{-3mm}
    \caption{Qualitative comparison on subject-driven generation tasks. More results are provided in \cref{sec:suppl-additonal-viz}.}
    \label{fig:compare-dreambooth}
    \vspace{-3mm}
\end{figure*}

\subsection{Main Results}

\noindent
\textbf{Subject driven generation.}
\cref{tab:compare-db} gives the results of weight-activation quantization on subject-driven generation task. It can be seen that the proposed method consistently outperforms other competitors under different bit-widths. For instance, the \MUL suppresses the IR-QLoRA by even 0.0428 DINO score on the W8A8 setup. Notably, the QLoRA and IR-QLoRA baselines, which use additional PTQ on the merged weights, suffer a significant performance drop under the W4A8 setup. In contrast, even the challenging log2-quantization of our \SHIFT works well under W4A8. We also give qualitative visualization in \cref{fig:compare-dreambooth}, where one can see that our \NAME can facilitate subject-faithful and photo-realistic image generation.

\begin{figure*}[!t]
    \centering
    \includegraphics[width=0.96\linewidth]{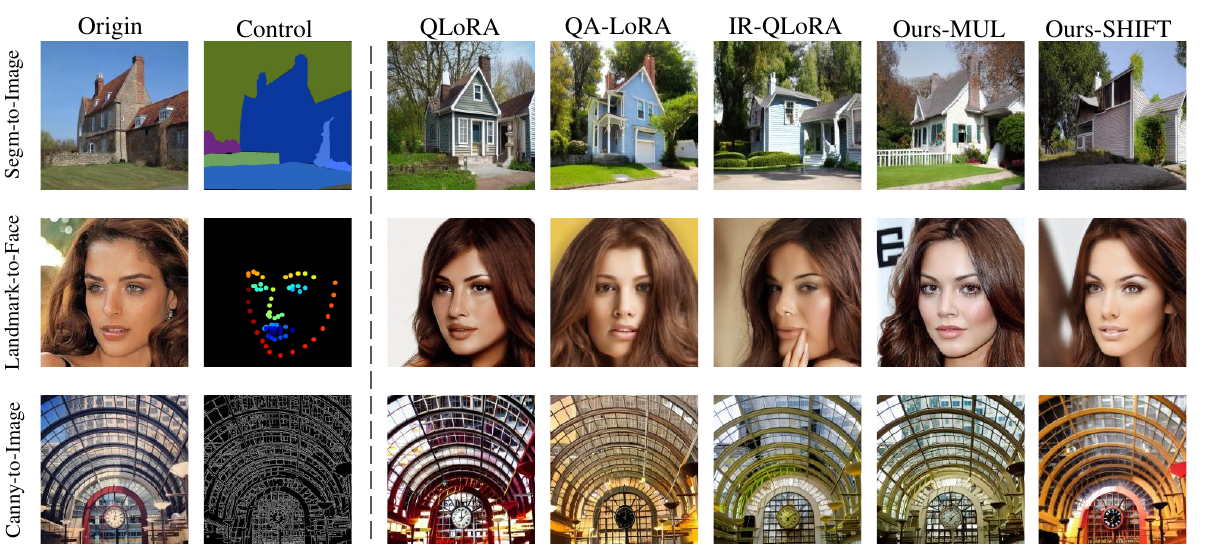}
    \vspace{-3mm}
    \caption{Qualitative comparison on controllable generation tasks. More results are provided in \cref{sec:suppl-additonal-viz}.}
    \label{fig:compare-control}
\end{figure*}

\noindent
\textbf{Controllable image generation.}
The results of controllable image generation are shown in \cref{tab:compare-control}. One can see that our IntLoRA continues to outperform existing strong baselines, \textit{e.g.}, our \MUL outperforms the IR-QLoRA by 4.96 FID on the 4-bit S2I setting. And it can be seen that QLoRA struggles to produce meaningful results at low 4 bit-width. We also give a qualitative comparison in \cref{fig:compare-control}, and it can be seen that the images generated by the IntLoRA-tuned model are well-matched with the control signals.

\begin{table}[!tb]
\centering
\caption{Quantitative comparison of FID$\downarrow$ score on controllable image generation.}
\vspace{-3mm}
\label{tab:compare-control}
\setlength{\tabcolsep}{5pt}
\scalebox{0.86}{
\begin{tabular}{@{}l|ccc|ccc@{}}
\toprule
\multirow{2}{*}{methods} & \multicolumn{3}{c|}{\textbf{8-bitwidth}} & \multicolumn{3}{c}{\textbf{4-bitwidth}}          \\
         & S2I   & L2F            & C2I   & S2I   & L2F    & C2I   \\ \midrule
LoRA(\FP) & 31.39 & 37.50          & 16.05 & 31.39 & 37.50  & 16.05 \\
\midrule
QLoRA    & 31.09 & 38.88          & 15.34 & 71.75 & 117.37 & 62.49 \\
QALoRA   & 31.32 & 38.88          & 15.34 & 31.51 & 43.09  & 16.73 \\
IR-QLoRA & 31.81 & 36.30          & 15.70 & 35.83 & 39.63  & 18.30 \\
\rowcolor[HTML]{EFEFEF} 
\SHIFT   & 31.38 & \textbf{34.46} & 15.76 & 32.85 & 35.06  & 17.65 \\
\rowcolor[HTML]{EFEFEF} 
\MUL                     & \textbf{31.08}  & 37.52 & \textbf{15.26} & \textbf{30.87} & \textbf{33.62} & \textbf{16.32} \\ \bottomrule
\end{tabular}%
}
\end{table}

\begin{table*}[!tb]
\centering
\caption{Comparison of training and inference efficiency with other methods. We fine-tuning the StableDiffusionV1.5 model on the Dreambooth task. The training speed is tested on one NVIDIA RTX 3090 GPU.}
\label{tab:efficiency-compare}
\vspace{-1mm}
\setlength{\tabcolsep}{10pt}
\scalebox{0.95}{
\begin{tabular}{@{}l|c|cc|ccc@{}}
\toprule
\multirow{2}{*}{method} & \multirow{2}{*}{nbits} & \multicolumn{2}{c|}{\textbf{Training Stage}} & \multicolumn{3}{c}{\textbf{Inference Stage}} \\
       &        & training speed & model size & PTQ & CLIP-I$\uparrow$ & CLIP-T$\uparrow$ \\ \midrule
LoRA~\citep{hu2021lora}   & W32A32 & 0.68s/img      & 7700MB     & \ding{52}            & 0.6968 & 0.2954 \\ \midrule
QLoRA~\citep{dettmers2024qlora}  & W8A8   & 0.85s/img      & 1925MB     & \ding{52}            & 0.6661 & 0.2824 \\
\rowcolor[HTML]{EFEFEF} 
\SHIFT (Ours)& W8A8   & 0.84s/img      & 1925MB     & \ding{56}   & 0.6842 & 0.2841 \\
\rowcolor[HTML]{EFEFEF} 
\MUL (Ours)  & W8A8   & 0.87s/img      & 1925MB     & \ding{56}   & 0.6882 & 0.2858 \\ \midrule
QLoRA~\citep{dettmers2024qlora}  & W4A8   & 0.85s/img      & 963.1MB      & \ding{52}            & 0.6134 & 0.2510 \\
\rowcolor[HTML]{EFEFEF} 
\SHIFT (Ours)& W4A8   & 0.84s/img      & 963.1MB      & \ding{56}             & 0.6716 & 0.2709 \\ 
\rowcolor[HTML]{EFEFEF} 
\MUL (Ours)   & W4A8   & 0.87s/img      & 963.1MB      & \ding{56}             & 0.6913 & 0.2710 \\
\bottomrule
\end{tabular}%
}
\end{table*}

\begin{figure*}[!t]
    \centering
    \includegraphics[width=0.96\linewidth]{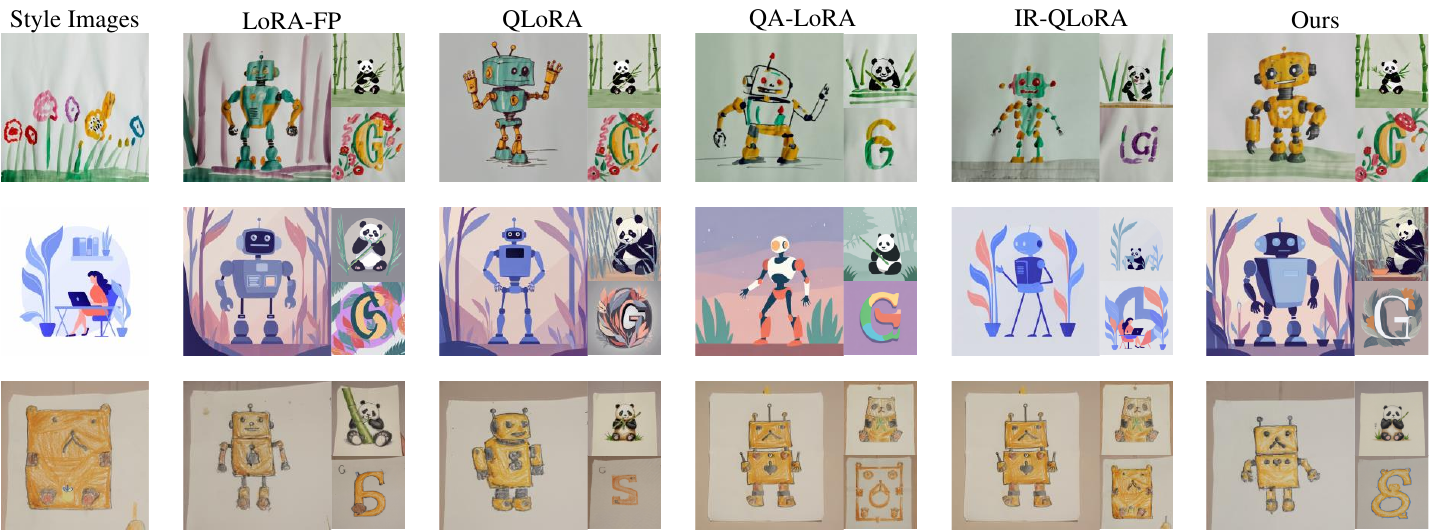}
    \vspace{-3mm}
    \caption{Qualitative comparison on style-customized generation. The text prompt is ``A friendly robot in [V] style",  ``A panda eating bamboo in [V] style", and ``The letter `G' in [V] style",  respectively. ``Ours'' denotes the \SHIFT. More results are provided in \cref{sec:suppl-additonal-viz}.}
    \label{fig:compare-style}
\end{figure*}

\noindent
\textbf{Style customized generation.}
The results of the style customized generation task are shown in \cref{fig:compare-style}. It can be seen that our \NAME achieves a favorable balance between style images and text prompts, whereas some existing approaches fail. For instance, in the third row, both the QALoRA and IR-QLoRA methods directly copy the original style image under the text prompt ``The letter `G’ in [V] style".

\subsection{Efficiency Comparison}

We compare the training and inference efficiency of our IntLoRA against other baselines in~\cref{tab:efficiency-compare}. As for training, both IntLoRA and QLoRA only need to load quantized pre-trained weights, which is more memory efficient compared to the vanilla LoRA fine-tuning. As a result, our IntLoRA and QLoRA have similar training speeds and memory costs. However, for the inference phase, our IntLoRA can naturally obtain the quantized merged weights without additional PTQ, thus streamlining the adaptation pipeline and avoiding potential performance degradation under low bit-width. In short, compared with existing methods which only focus on training efficiency, our IntLoRA presents a both training and inference efficient paradigm.

\subsection{Ablation Studies}
\label{sec:ablation}

\begin{figure}[!t]
    \centering
    \includegraphics[width=0.84\linewidth]{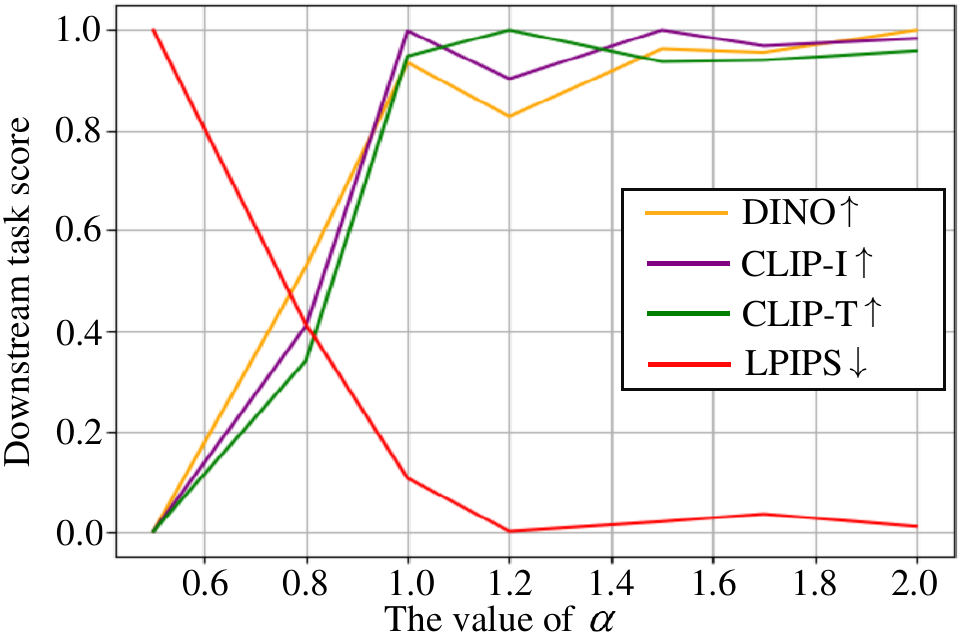}
    \vspace{-3mm}
    \caption{\small{The normalized performance under different $\alpha$.}}
    \label{fig:ablation-smooth-factor}
    \vspace{-3mm}
\end{figure}

\noindent
\textbf{Ablation on the smoothing factor.}
As discussed in~\cref{sec:method}, there is a dilemma in choosing an appropriate $\sigma_\mathbf{R}$. For example, setting it too large can lead to information loss of the original weights, while a too-small one results in a large quantization error. To this end, we introduce $r^{\alpha}$ in the proposed VMC as the hyperparameter to search for the task-oriented variance. We give the downstream task performance with varying $\alpha$ in~\cref{fig:ablation-smooth-factor}. It can be seen that setting $\sigma_\mathbf{R}$ slightly smaller than $\sigma_\mathbf{W}$ can obtain better performance, indicating that the information loss has a greater impact than the quantization error. In the implementation, we chose a moderate smoothing factor $\alpha=1.5$ for the trade-off.

\noindent
\textbf{Distribution selection for auxiliary matrix.}
In this work, the auxiliary matrix $\mathbf{R}$ plays a crucial role in both AQS and VMC. Therefore, the distribution shape of $\mathbf{R}$ can potentially influence the performance. To this end, we investigate different distribution shapes of $\mathbf{R}$ through ablation experiments and give the results in ~\cref{tab:ablation-distribution-choice}. It can be seen that the Laplace distribution performs better than other options on most metrics. This is because a light-tailed distribution, such as Laplace, clusters most samples around zero, which facilitates smaller errors for log2-quantization. Therefore, the light-tailed distribution performs empirically better than its heavy-tailed counterparts.

\begin{table}[!tb]
    \centering
    \caption{Ablation on different distribution shape choices of the auxiliary matrix.}
    \vspace{-3mm}
    \label{tab:ablation-distribution-choice}
    \setlength{\tabcolsep}{9pt}
    \scalebox{0.88}{
    \begin{tabular}{@{}l|cccc@{}}
    \toprule
    settings & DINO$\uparrow$   & CLIP-I$\uparrow$ & CLIP-T$\uparrow$ & LPIPS$\downarrow$  \\ \midrule
    Guassian & 0.4135 & 0.6756 & 0.2492 & 0.8179 \\
    Cauchy   & 0.1367 & 0.5617 & 0.1870 & 0.8067 \\
    StudentT & 0.2935 & 0.6420 & 0.2487 & 0.8048 \\
    Laplace  & 0.4492 & 0.6980 & 0.2588 & 0.8110 \\ \bottomrule
    \end{tabular}%
    }
    \vspace{-3mm}
\end{table}

\section{Further Discussion}

\noindent
\textbf{Results on NLP tasks.}
In addition to fine-tuning diffusion models for image generation, we further validate the effectiveness of the NLP tasks. Specifically, we fine-tune the Llama3-8B model~\citep{dubey2024llama3} and use the MetaMathQA dataset~\citep{yu2023metamath} for
training and GSM8K dataset~\citep{cobbe2021gsmk} for testing. The comparison results are shown in~\cref{tab:mathqa-compare}. It can be seen that our \NAME maintains stable performance when transferring to natural language. For example, \MUL outperforms QLoRA by 0.17\% QA accuracy under 8-bit quantization, demonstrating the generalization of our IntLoRA.

\noindent
\textbf{Difference from EfficientDM.}
EfficientDM~\citep{he2023efficientdm} employs QAT-like LoRA fine-tuning on the \FP diffusion weights for network quantization. Despite it is inference-efficient as it can directly produce quantized merged weights, we would like to point out that it is not training-efficient. Specifically, EfficientDM requires load \FP pre-trained weights at the training stage, which is unacceptable for fine-tuning large-size models on consumer-level GPUs. By contrast, our IntLoRA is both training and inference efficient. 
Moreover, we also explore our IntLoRA on the diffusion quantization task. The results are shown in~\cref{tab:compare-quant-efficientdm}. It can be seen that our \MUL achieves even better performance than the EfficientDM. It should be noted that our IntLoRA only needs to load the quantized weights during calibration instead of the floating-point weights in EfficientDM, thus reducing the training memory cost.

\begin{table}[!tb]
\centering
\caption{Comparison on the natural language task of mathematical answering. More qualitative results in~\cref{sec:suppl-additonal-viz}.}
\label{tab:mathqa-compare}
\vspace{-3mm}
\setlength{\tabcolsep}{1.8pt}
\scalebox{0.9}{
\begin{tabular}{@{}l|cccccc@{}}
\toprule
Methods  & LoRA & QLoRA   & QA-LoRA  & \cellcolor[HTML]{EFEFEF}{Ours$\rm{_{SHIFT}}$}    & \cellcolor[HTML]{EFEFEF}{Ours$\rm{_{MUL}}$}  \\ \midrule nbits
 & W16A16   &   W8A8 & W8A8 & \cellcolor[HTML]{EFEFEF}{W8A8}   & \cellcolor[HTML]{EFEFEF}{W8A8}   \\
accuracy & 64.24\% & 64.06\% & 63.53\% & \cellcolor[HTML]{EFEFEF}{64.10\%} & \cellcolor[HTML]{EFEFEF}{\textbf{64.23\%}} \\ \bottomrule
\end{tabular}%
}
\end{table}

\begin{table}[!tb]
\centering
\caption{Comparison with EfficientDM on W4A4 diffusion model quantization. We evaluate on the ImageNet $256 \times
256$ image generation, and train with ddim\_step=20 on LDM-4 model with 500 training epochs.}
\vspace{-3mm}
\label{tab:compare-quant-efficientdm}
\setlength{\tabcolsep}{8pt}
\scalebox{0.88}{
\begin{tabular}{@{}l|ccccc@{}}
\toprule
methods   & IS$\uparrow$     & FID$\downarrow$    & sFID$\downarrow$  & precision$\uparrow$    \\ \midrule
EfficientDM    & 178.20 & 13.42 & 26.67 & 0.70        \\
\rowcolor[HTML]{EFEFEF} 
Ours$\rm{_{SHIFT}}$    & 116.50 & 20.20 & 26.79 & 0.63        \\
\rowcolor[HTML]{EFEFEF} 
Ours$\rm{_{MUL}}$     & \textbf{199.20} & \textbf{10.43} & \textbf{24.02} & \textbf{0.79}   \\ \bottomrule
\end{tabular}%
}
\end{table}

\section{Conclusion}
We propose \NAME, which employs integer-type low-rank parameters, to remove the additional PTQ on the merged weights. Specifically, we introduce the quantization-adaptation separation to allow the coexistence of zero-initialized gradient and quantization-friendly distribution. We further develop the multiplicative low-rank adaptation to achieve a decoupled quantizer of pre-trained and adaptation weights, accompanied by the variance matching control to adjust the variance for accurate adaptation control. Benefiting from these elegant designs, we provide two variants of \NAME, which either use int-multiplication or bit-shifting to adapt the quantized pre-trained models. Through transferring the adaptation weights to the integer arithmetic, our \NAME demonstrates its effectiveness across different pre-trained models and various downstream tasks, while exhibiting impressive both training and inference efficiency.

\section*{Acknowledges}
This work is supported in part by the National Natural Science Foundation of China, under Grant (62302309, 62171248), Shenzhen Science and Technology Program (JCYJ20220818101014030, JCYJ20220818101012025).

\section*{Impact Statement}
This work aims to achieve efficient fine-tuning of quantized diffusion models, reducing both training and inference costs without sacrificing performance. It has no ethical concerns and can lower the resource barrier for model customization, thus enabling broader access to generative AI.

\nocite{langley00}

\bibliography{icml}
\bibliographystyle{icml2025}

\clearpage
\newpage
\appendix
\onecolumn

\section{Summery of IntLoRA Algorithm}
\label{sec:suppl-algo}

Before tuning, we pre-process the pre-trained weights in~\cref{algo:preprocess}, followed by the forward process of \MUL and \SHIFT in~\cref{algo:forward-mul} and~\cref{algo:forward-shift}, respectively.

\begin{algorithm}[!h]
\caption{The weight pre-process of the linear layer in IntLoRA} 
\label{algo:preprocess}
\begin{algorithmic}
\STATE {\bfseries Input:} 
Pre-trained wight $\mathbf{W} \in \mathbb{R}^{C_{out}\times C_{in}}$, auxiliary matrix $\mathbf{R}\in \mathbb{R}^{C_{out}\times C_{in}}$, smooth factor $\alpha \in \mathbb{R}$
\STATE {\bfseries Output:} Quantitzed pre-trained weights $\mathbf{W}_\mathrm{round}$, scaling factor $s_\mathrm{round}$, zero point $z_\mathrm{round}$
\STATE 
sigma\_R $\gets$ variance estimation of $\mathbf{R}$
\STATE sigma\_W $\gets$ variance estimation of $\mathbf{W}$
\STATE r = (sigma\_W / sigma\_R)$^\alpha$
\STATE $\mathbf{R}_\mathrm{star}$ = r * $\mathbf{R}$
\STATE $\mathbf{W}_\mathrm{process} = \mathbf{W-R_\mathrm{star}}$ 
\STATE $\mathbf{W}_\mathrm{round}, s_\mathrm{round}, z_\mathrm{round} \gets$ uniform\_quantizer($\mathbf{W}_\mathrm{process}$)
\end{algorithmic}
\end{algorithm}

\begin{algorithm}[!h]
\caption{The forward process of the linear layer in \MUL} 
\label{algo:forward-mul}
\begin{algorithmic}
\STATE {\bfseries Input:} 
Pre-processed quantized weights  $\mathbf{W}_\mathrm{round}$, scaling factor $s_\mathrm{round}$, zero point $z_\mathrm{round}$, auxiliary matrix $\mathbf{R}_\mathrm{star}\in \mathbb{R}^{C_{out}\times C_{in}}$, LoRA parameters $\mathbf{A} \in \mathbb{R}^{C_{out}\times d}$,  $\mathbf{B} \in \mathbb{R}^{C_{out}\times d}$, input tensor $\mathbf{x}\in \mathbb{R}^{C_{in}\times L}$
\STATE {\bfseries Output:} Output tensor $\mathbf{y} \in \mathbb{R}^{C_{out}\times L}$
\STATE 
$\mathbf{W}_\mathrm{adapt} = s_\mathrm{round} \cdot \mathbf{I} + \frac{1}{\mathbf{W}_\mathrm{round}-z_\mathrm{round}}\odot(\mathbf{R_\mathrm{star}+AB})$
\STATE
$\mathbf{W}_\mathrm{adapt}^\texttt{INT}, s_\mathrm{adapt}, z_\mathrm{adapt} \gets \text{uniform\_quantizer}(\mathbf{W}_\mathrm{adapt})$
\STATE
$\mathbf{x}^\texttt{INT}, s_\mathrm{x} \gets \text{act\_quantizer}(\mathbf{x})$
\STATE
$\mathbf{W}_\mathrm{merge} = (\mathbf{W}_\mathrm{adapt}^\texttt{INT} -z_\mathrm{adapt})\odot (\mathbf{W}_\mathrm{round}-z_\mathrm{round})$
\STATE
$\mathbf{y}=s_\mathrm{x} s_\mathrm{round} \mathbf{W}_\mathrm{merge}\mathbf{x}^\texttt{INT}$
\end{algorithmic}
\end{algorithm}

\begin{algorithm}[!h]
\caption{The forward process of the linear layer in \SHIFT} 
\label{algo:forward-shift}
\begin{algorithmic}
\STATE {\bfseries Input:} 
Pre-processed quantized weights  $\mathbf{W}_\mathrm{round}$, scaling factor $s_\mathrm{round}$, zero point $z_\mathrm{round}$, auxiliary matrix $\mathbf{R}_\mathrm{star}\in \mathbb{R}^{C_{out}\times C_{in}}$, LoRA parameters $\mathbf{A} \in \mathbb{R}^{C_{out}\times d}$,  $\mathbf{B} \in \mathbb{R}^{C_{out}\times d}$, input tensor $\mathbf{x}\in \mathbb{R}^{C_{in}\times L}$, desired bit-width $b$, pre-defined max bit-width number $N = 32$
\STATE {\bfseries Output:} Output tensor $\mathbf{y} \in \mathbb{R}^{C_{out}\times L}$
\STATE 
$\mathbf{W}_\mathrm{adapt} = s_\mathrm{round} \cdot \mathbf{I} + \frac{1}{\mathbf{W}_\mathrm{round}-z_\mathrm{round}}\odot(\mathbf{R_\mathrm{star}+AB})$
\STATE
shift = $\text{clip}(\lfloor-\log_2|\mathbf{W}_\mathrm{adapt}|\rceil,0,2^b -1)$
\STATE
$\mathbf{W}_\mathrm{merge} = \frac{1}{2^N} \odot \sign(\mathbf{W}_\mathrm{adapt}) \odot [(\mathbf{W}_\mathrm{round}-z)\ll(N-\mathrm{shift})]$
\STATE
$\mathbf{x}^\texttt{INT}, s_\mathrm{x} \gets \text{act\_quantizer}(\mathbf{x})$
\STATE
$\mathbf{y}=s_\mathrm{x} \mathbf{W}_\mathrm{merge}\mathbf{x}^\texttt{INT}$
\end{algorithmic}
\end{algorithm}

\section{Distribution Visualization of $\sigma_\mathbf{R}$}
\begin{figure*}[!t]
    \centering
    \includegraphics[width=0.9\linewidth]{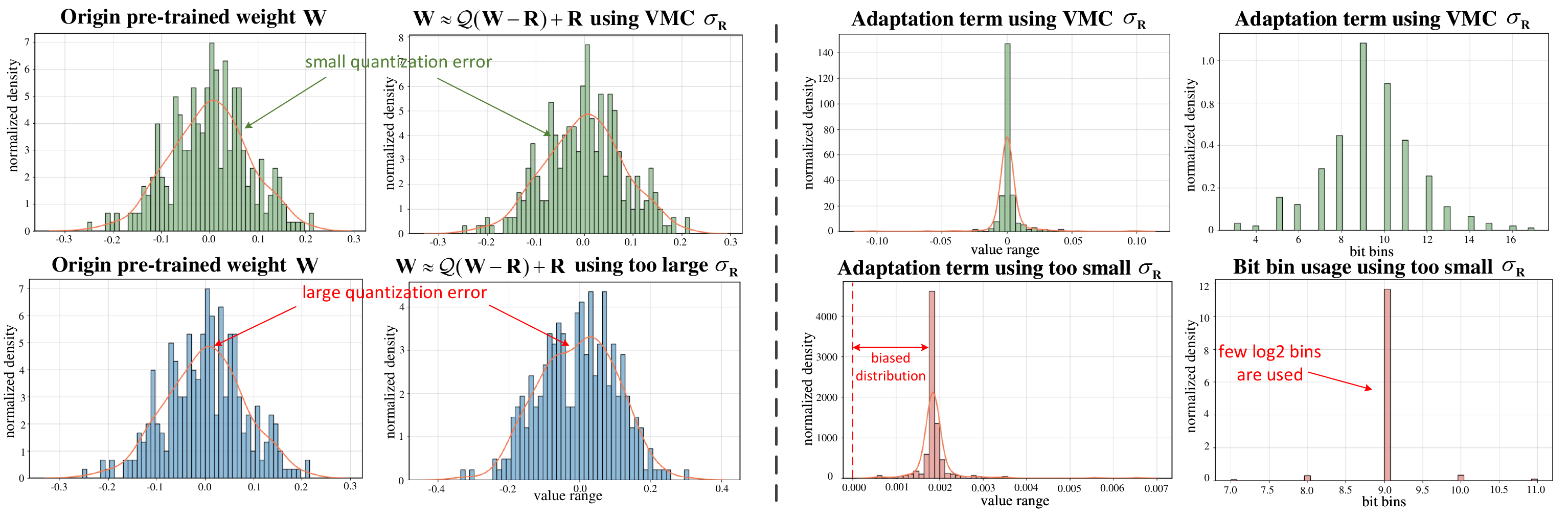}
    \vspace{-3mm}
    \caption{The distribution visualization using Kernel Density Estimate (KDE) on different weight tensors. \textbf{Left}: the KDE plot of pre-trained weights and estimated weights under different $\sigma_\mathbf{R}$. \textbf{Right}: the KDE plot of the adaptation term and the log2 bins usage with different $\sigma_\mathbf{R}$.}
    \label{fig:distribution_viz_ablation}
\end{figure*}

In \cref{sec:implementation}, we have theoretically pointed out that there is a choice dilemma for $\sigma_\mathbf{R}$. Here we elaborate on its effect through distribution visualization. Specifically, we remove the VMC and use a scaling scalar to generate a too-large or too-small auxiliary variance, followed by the log2 quantization on the adaptation term. The results are shown in~\cref{fig:distribution_viz_ablation}.
On the one hand, setting $\sigma_\mathbf{R}$ too large can lead to a low correlation $\rho(\mathbf{W},\mathbf{W-R})$, which makes it hard to reconstruct $\mathbf{W}$ from $\mathbf{W-R}$ using estimator $\mathbf{W} \approx \mathcal{Q}(\mathbf{W-R})+\mathbf{R}$. On the other hand, a too small $\sigma_\mathbf{R}$ prevents the expectation of adaptation term converging to zero, causing few log bins to be used. In experiments, we find that the training of both settings fails to converge. By contrast, the proposed VMC can precisely control  $\sigma_\mathbf{R}$ to allow most values of the adaptation term to be zero-neighbored, facilitating more challenging log2 quantization.  Moreover, it should be noted that the too-small $\sigma_\mathbf{R}$ can also be regarded as an approximation of direct quantization on the zero-initialized $\mathbf{AB}$, and thus the experimental results also justify the AQS for zero-initialized $\mathbf{AB}$.

\section{More Implementation Details}
\label{sec:suppl-additoinal-details}
For the subject-driven generation, we use the AdamW optimizer with a weight decay of 1e-2 and fine-tune the query, key, value, and output projection layer.  The learning rate is set to 6e-5. The batch size is set to 1, and the number of training steps is 400. The rank of the LoRA is set to 4. We adopt the prior preservation strategy as Dreambooth~\citep{ruiz2023dreambooth} to generate 200 class images. 
For the controllable generation, we fine-tune the model for 11 epochs for Canny-to-Image tasks and 20 epochs for Landmark-to-Face and Segmentation-to-Image tasks. The learning rate is set to 1e-5 using the AdamW optimizer. The LoRA rank is set to 4. The batch size is set to 8 and the image resolution is $512 \times 512$ for all three tasks. 
For the style-customized generation, we fine-tune the pre-trained model using the AdamW optimizer with a learning rate of 5e-5. Since it involves a larger SDXL, we chose a relatively large LoRA rank of 64 for all compared methods, since there is only one style image as well as the larger pre-trained parameters. We fine-tune for 500 steps with batch size 1. Similar to StyleDrop~\citep{sohn2023styledrop}, we only use one image as the style image and find it works well. The style images and text prompts for evaluation are given in~\cref{sec:suppl-additonal-viz}. The variance ratio in the variance matching control is surrogated as the value range ratio, \textit{i.e.}, $r = \frac{\max\{|\max(\mathbf{W})|,|\min(\mathbf{W})|\}}{\min\{|\max(\mathbf{R})|,|\min(\mathbf{R})|\}}$.
We append the trainable low-rank parameters on the Query, Key, Value, and Out projection, in the attention layers~\citep{vaswani2017attention} and keep all other layers frozen and quantized. 
The rank of LoRA is set to 4 for subject-driven generation and controllable generation, and 64 for style-customized generation.

\section{Additional Ablation Experiments}

\begin{figure}[!t]
    \centering
    \includegraphics[width=0.98\linewidth]{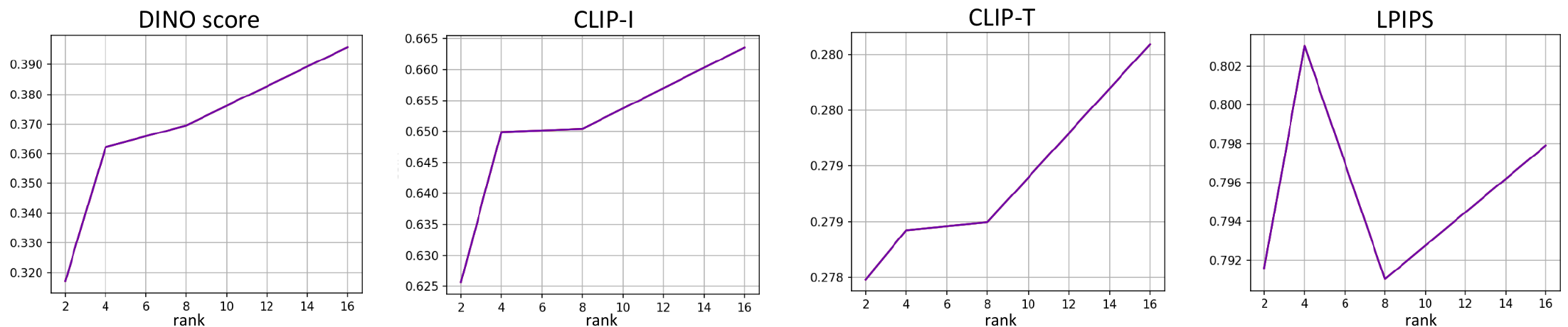}
    \caption{Ablation experiments of different LoRA ranks.}
    \label{fig:suppl-ablation-lora-rank}
\end{figure}

\noindent
\textbf{Ablation on the LoRA rank.}
The low-rank $d$ in LoRA is a trade-off between performance and efficiency. A larger rank improves the adaptation ability by training more parameters but comes with larger training and storage costs, and vice versa. Here, we give the impact of different rank setups on performance in ~\cref{fig:suppl-ablation-lora-rank}. One can see that the performance generally improves as we increase the rank, but the rate of growth varies. For instance, the increased speed from rank=4 to rank=8 increases inferior to the one from rank=2 to rank=4. Moreover, increasing the rank to 16 can generally obtain better results than its lower counterpart. In practice, considering the trade-off between performance and efficiency, we select a moderate rate rank=4.

\begin{wrapfigure}{r}{0.35\linewidth}
    \centering
    \vspace{-3mm}
    \includegraphics[width=0.85\linewidth]{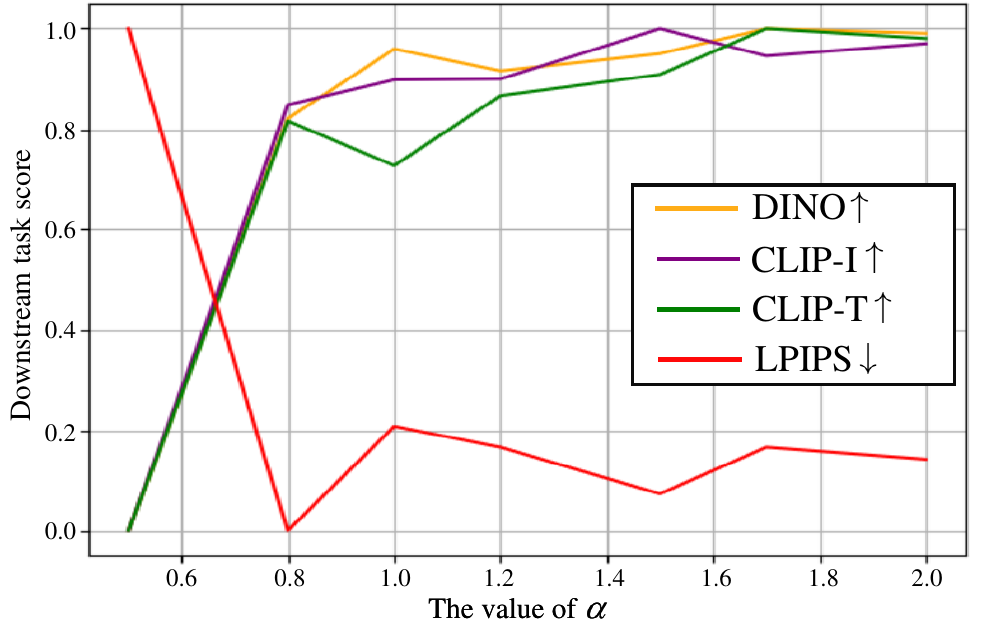}
    \vspace{-2mm}
    \caption{The effects of VMC for \MUL.}
    \label{fig:suppl-vmc-for-intmul}
    \vspace{-3mm}
\end{wrapfigure}

\noindent
\textbf{The effects of variance matching control for \MUL.}
In this work, we propose the variance matching control to adjust the variance of $\mathbf{R}$, so that allows the log2-quantization of the adaptation term to obtain the \SHIFT. In other words, the VMC is primarily introduced for \SHIFT. Despite we also apply the VMC to \MUL, given the \MUL does not require such strict constraints on the distribution shape of the adaptation term, it is interesting to investigate the influence of variance matching control on the performance of \MUL. To this end, we adjust the smoothing factor $\alpha$ to adjust the strength of the VMC, \textit{e.g.}, setting $\alpha$ to zero can lead to the removal of the VMC. The results of the \MUL under different VMC scales are shown in ~\cref{fig:suppl-vmc-for-intmul}. As one can see, despite the VMC being initially proposed for the log2-quantization, the well-structured distribution also facilitates uniform quantization. For example, when we set the $\alpha$ approaching zero, \textit{i.e.}, the VMC is close to being removed, and the performance of \MUL appears similar pattern as the \SHIFT, which suffers a significant performance drop. Moreover, the performance gains gradually converge when the $\alpha > 1.5$. In short, the VMC can not only allow the log2-quantization to work but also improve the performance of the uniform quantization.

\begin{figure*}[!t]
    \centering
    \includegraphics[width=1.\linewidth]{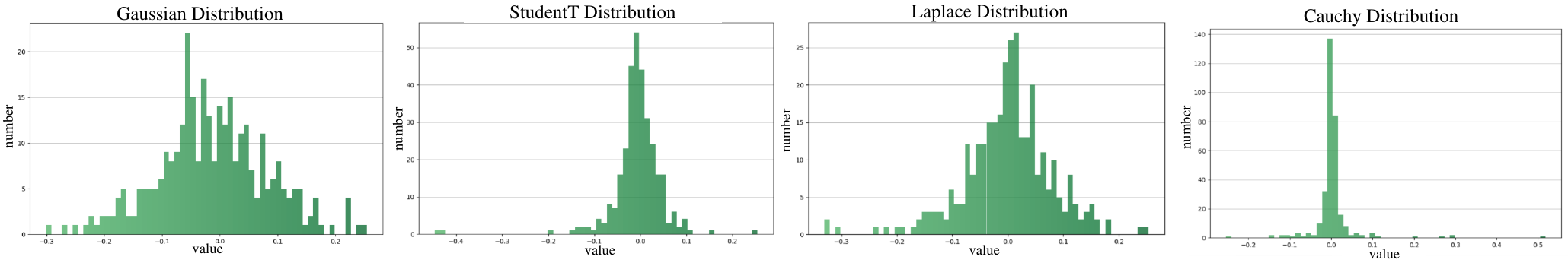}
    \vspace{-6mm}
    \caption{The shape of different distributions for initialing the auxiliary matrix.}
    \label{fig:suppl-distribution-aux}
\end{figure*}

\noindent
\textbf{Distribution shape for auxiliary matrix.}
In~\cref{sec:ablation}, we provided different symmetric distributions including Gaussian, StudentT, Laplace, and Cauchy. ~\cref{fig:suppl-distribution-aux} gives the results of sampling from different distributions. The Laplace distribution possesses light tails, and the shape of the distribution is convex, \textit{i.e.}, $f''(x)>0, x\neq0$. This unique property makes it easy to control the value of the adaptation term to produce distributions that are friendly to log2 quantization, \textit{i.e.},  most samples are clustered around the zero to use as many bins as possible. This analysis is also verified by the experiments in ~\cref{tab:ablation-distribution-choice}, which shows that the Laplace distribution achieves the best performance.

\begin{figure*}[!t]
    \centering
    \includegraphics[width=1.\linewidth]{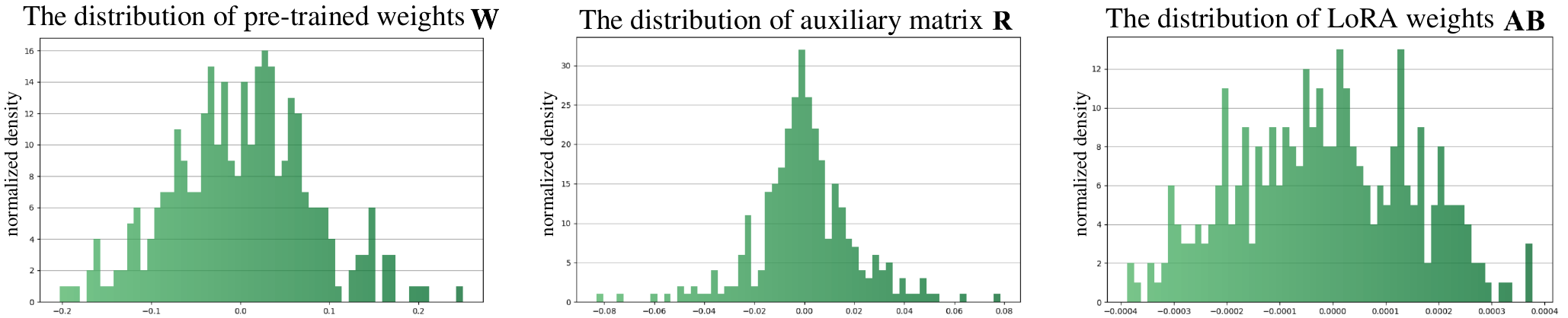}
    \vspace{-6mm}
    \caption{ {The distribution visualization of the original weights $\mathbf{W}$, the auxiliary matrix $\mathbf{R}$, and the learned low-rank weights $\mathbf{AB}$.}}
    \label{fig:suppl-W-R-AB}
\end{figure*}

{\section{Impacts from the Auxiliary Matrix}
In~\cref{eq:quant-peft-separate} of the proposed AQS, we introduce an additional auxiliary matrix $\mathbf{R}$ to the original pre-trained weight $\mathbf{W}_0$ to achieve adaptation-quantization separation. However, this extra $\mathbf{R}$ potentially introduces outliers and thus causes quantization error for $\mathbf{W}$. Here, we point out that since the proposed VMC can control the range of $\mathbf{R}$ through the variance scaling factor $r = \sigma_\mathbf{W}/\sigma_\mathbf{R}$, the introduction of $\mathbf{R}$ in the AQS is ensured not result in additional outliers. For validation, we also give the distribution visualization of the original $\mathbf{W}$ and the VMC re-scaled $\mathbf{R}$ in~\cref{fig:suppl-W-R-AB}. It can be seen that the range of $\mathbf{R}$ is effectively controlled within the range of $\mathbf{W}$, thus effectively avoiding the detrimental effect of additional outliers.}

\begin{wrapfigure}{r}{0.35\linewidth}
    \centering
    \includegraphics[width=0.7\linewidth]{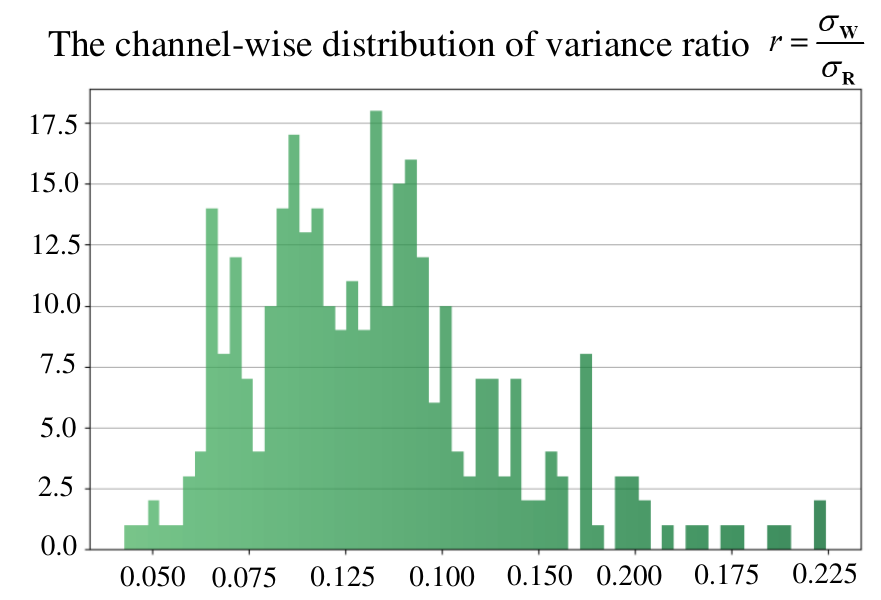}
    \vspace{-3mm}
    \caption{The value distribution of the channel-wise variance ratio $r$}
    \label{fig:suppl-distribution-r}
\end{wrapfigure}

\section{Justification for the Value Orders}
\label{sec:suppl-ab-less-than-R}
A key assumption in the derivation for VMC is that the learned values of low-rank parameters $\mathbf{AB}$ are orders of magnitude smaller than the auxiliary matrix $\mathbf{R}$. Based on this assumption we ignore $\mathbf{AB}$ as an approximation. Here, we give the specific evidence for this approximation. Specifically, we visualize the weights of the trained $\mathbf{AB}$ and the distribution of $\mathbf{R}$, as shown in~\cref{fig:suppl-W-R-AB}. It can be seen the range of $\mathbf{AB}$ is constrained to $[-0.0004,0.0004]$, while the range of $\mathbf{R}$ is $[-0.08, 008]$. Therefore, the experimental visualization above confirms the soundness of our approximation. Since the $\mathbf{AB}$ in LoRA is zero-initialized, it tends to be distributed around zero with learned small values aiming to not disturb the pre-training weights too much.

\section{Limitation and Future Works}
Although the proposed IntLoRA can effectively improve the efficiency of diffusion model fine-tuning by allowing the adaptation parameters also on the integer arithmetic, the proposed framework can be further improved in the following aspects. First, although the trainable low-rank weights are quantized with STE, these quantized weights are still in \FP type during tuning to enable accurate gradient updates. Therefore, it is promising to specifically design integer-type propagation. Despite this seems challenging, it can further reduce the training cost and accelerate the adaptation process. Second, although we introduce a feasible way that uses hyperparameter search of the smoothing factor $\alpha$ to find a compatible $\sigma_\mathbf{R}$ as well as the appropriate distribution shape of $\mathbf{R}$, it can be more elegant if we can use advanced mathematical analysis techniques, such as functional analysis, to find the statistical properties a suitable $\mathbf{R}$ should satisfy. Third, this work mainly focuses on the efficient acceleration of LoRA due to its prevalence among the PEFT techniques. Applications to other  PEFT methods for hardware-efficient adaptation could be interesting future work.

\section{Additional Visualization Results}
\label{sec:suppl-additonal-viz}

\begin{figure*}[p]
    \centering
    \includegraphics[width=0.9\linewidth]{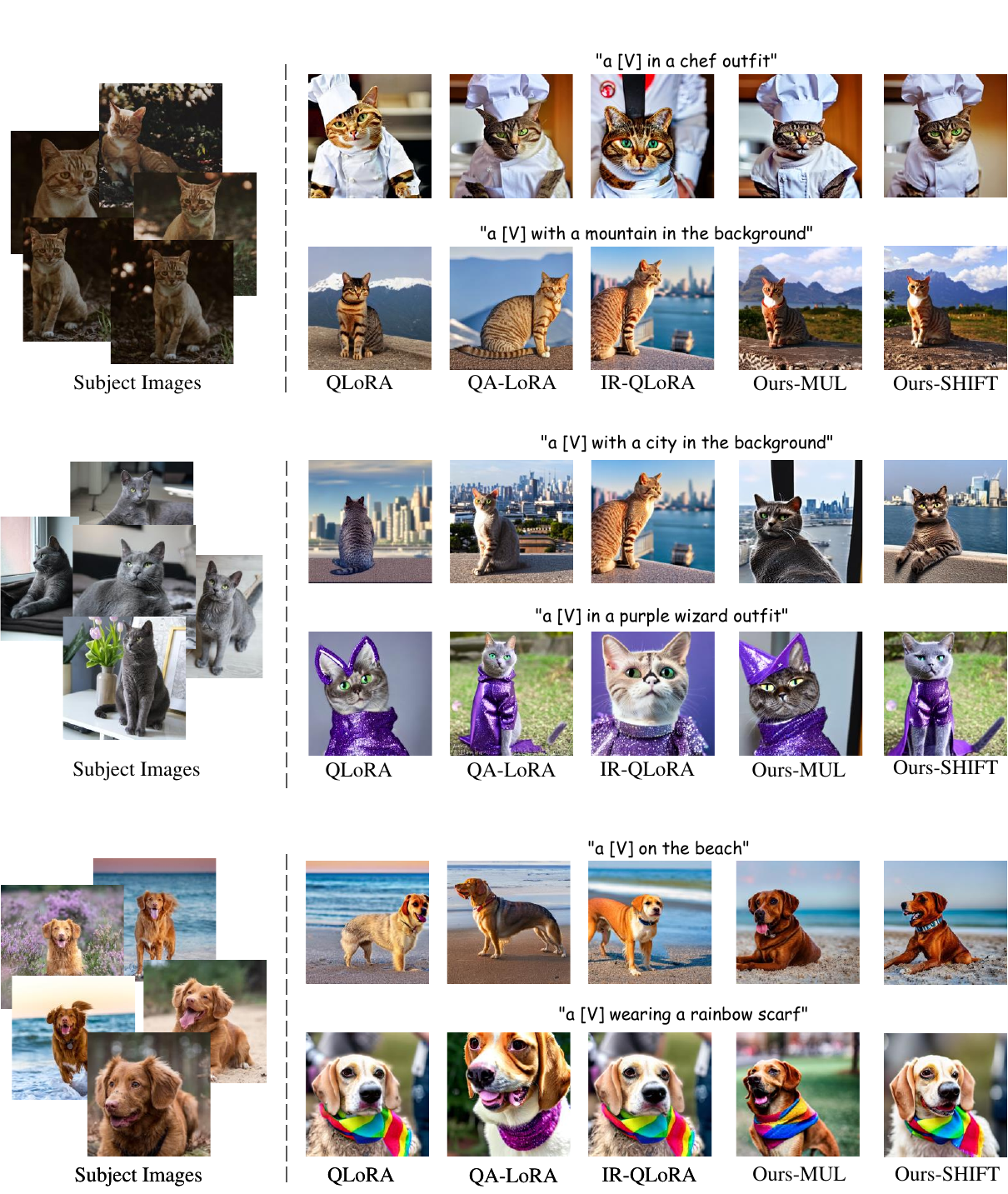}
    \caption{More qualitative comparison results on subject-driven generation.}
    \label{fig:suppl-dreambooth}
\end{figure*}

\begin{figure*}[p]
    \centering
    \includegraphics[width=0.9\linewidth]{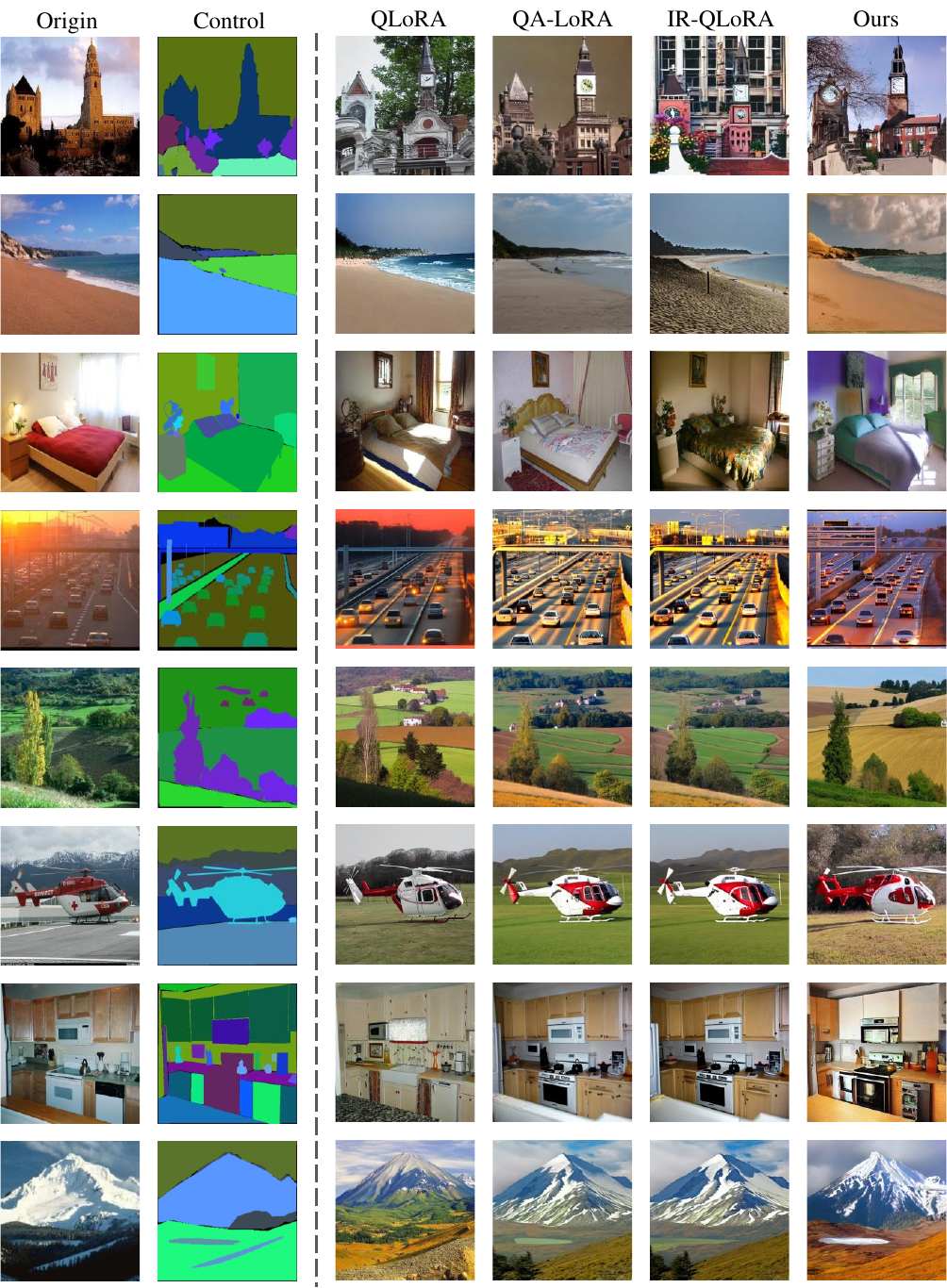}
    \caption{More qualitative comparison results on segmentation to image task. The `Ours' denotes the \SHIFT. Zoom in for better effects.}
    \label{fig:suppl-control-segm}
\end{figure*}

\begin{figure*}[p]
    \centering
    \includegraphics[width=0.9\linewidth]{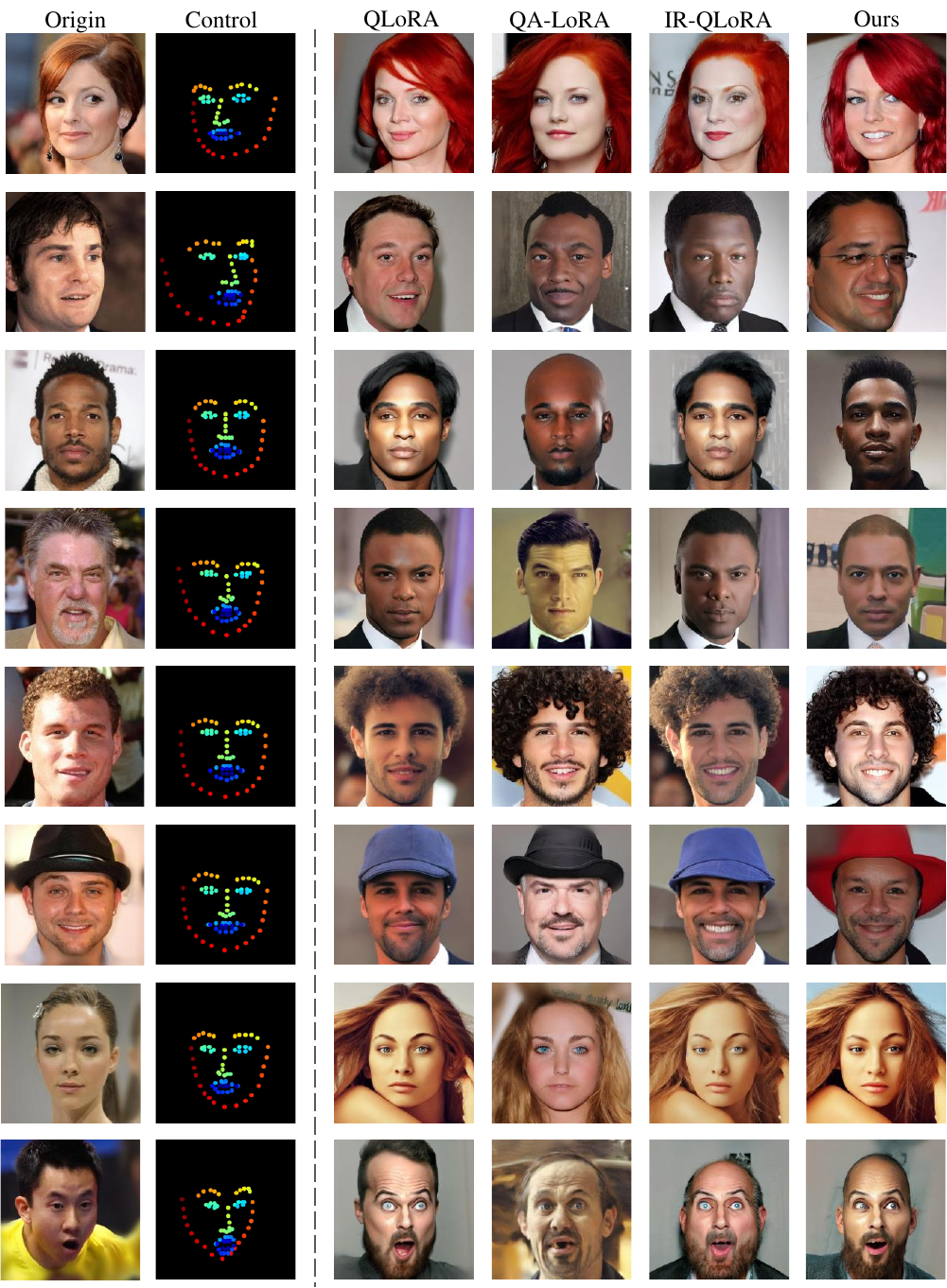}
    \caption{More qualitative comparison results on landmark to face task. The `Ours' denotes the \SHIFT. Zoom in for better effects.}
    \label{fig:suppl-control-landmark}
\end{figure*}

\begin{figure*}[p]
    \centering
    \includegraphics[width=0.9\linewidth]{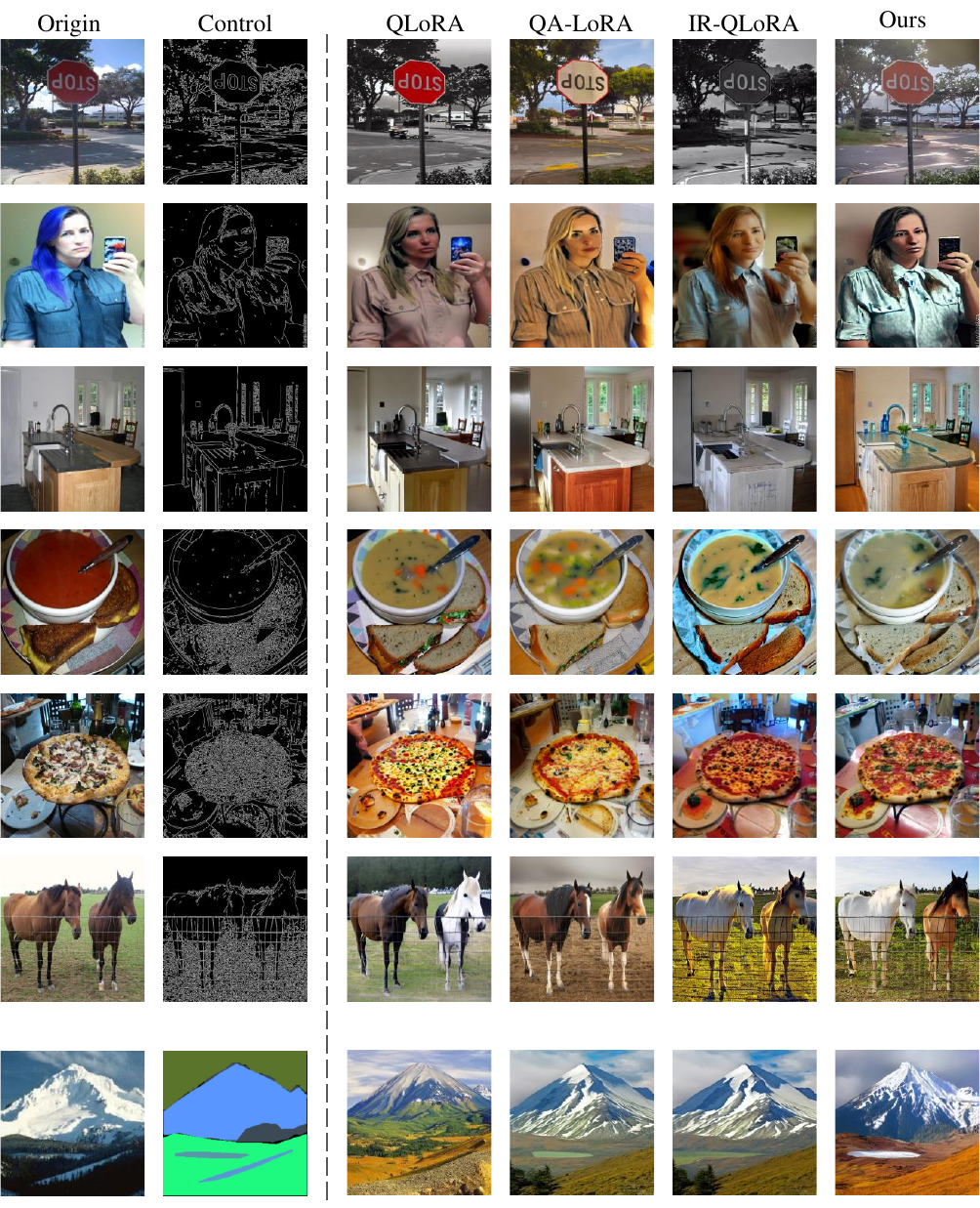}
    \caption{More qualitative comparison results on canny to image task. The `Ours' denotes the \SHIFT. Zoom in for better effects.}
    \label{fig:suppl-control-canny}
\end{figure*}

\begin{figure*}[p]
    \centering
    \includegraphics[width=0.9\linewidth]{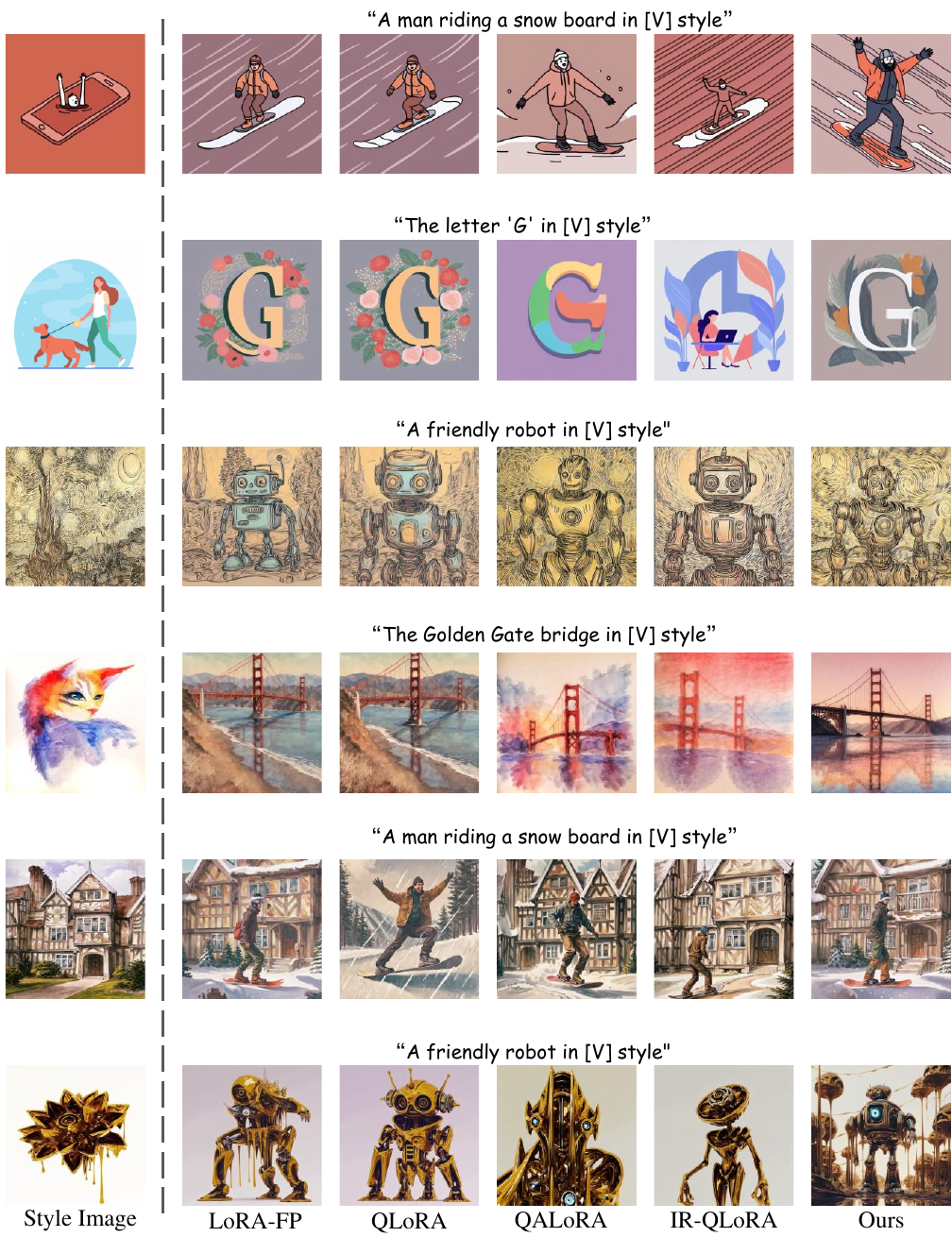}
    \caption{More qualitative comparison results on style-accustomed generation. The `Ours' denotes the \SHIFT. Zoom in for better effects.}
    \label{fig:suppl-style}
\end{figure*}

\begin{figure*}[p]
    \centering
    \includegraphics[width=0.9\linewidth]{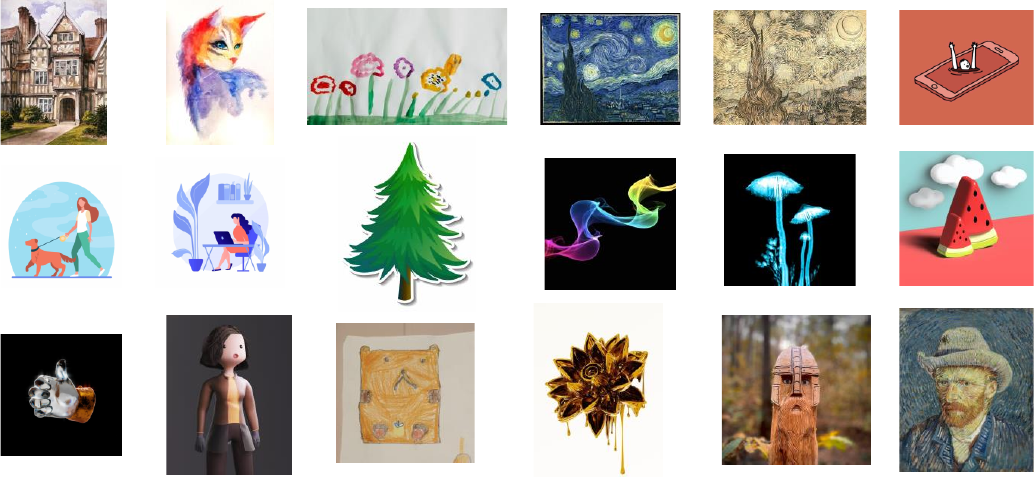}
    \caption{The style images used for training on the style-customized generation task. The text prompts are ``A friendly robot in [V] style", ``A fluffy baby sloth with a knitted hat trying to figure out a laptop in [V] style", ``A panda eating bamboo in [V] style", ``The Golden Gate bridge in [V] style", ``A man riding a snowboard in [V] style", ``The letter `G' in [V] style". }
    \label{fig:suppl-styledrop-setup}
\end{figure*}

\begin{itemize}
    \item \cref{fig:suppl-dreambooth} gives more samples on the subject-driven generation tasks.
    \item In \cref{fig:suppl-control-segm}, we give more samples of the segmentation-to-image tasks.
    \item In \cref{fig:suppl-control-landmark}, we give more samples of the face landmark-to-face image tasks.
    \item In \cref{fig:suppl-control-canny}, we give more samples of the canny edge-to-image tasks.
    \item \cref{fig:suppl-style} provides more samples of the results of the style-accustomed generation. 
    \item In \cref{fig:suppl-styledrop-setup}, we give the style images and the text prompts used for evaluation on the style customized generation tasks.
    \item In \cref{box:suppl-case-study}, we give some case studies of the mathematical question-answering task using the fine-tuned Llama3-8B model.
    
\end{itemize}

\clearpage

\clearpage

\clearpage

\clearpage

\clearpage

\clearpage

\clearpage
\subsection{Cases of mathematical question answering.}
\label{fig:suppl-mathQA-viz}
\begin{tcolorbox}[title={\large Case Study 1}, colback=cgray, colframe=cGreen, colbacktitle=cGreen, coltitle=white]

\textbf{Question}: 
Janet’s ducks lay 16 eggs per day. She eats three for breakfast every morning and bakes muffins for her friends every day with four. She sells the remainder at the farmers' market daily for \$2 per fresh duck egg. How much in dollars does she make every day at the farmers' market?
\textcolor{cGreen}{\textbf{(Ground-truth answer is 18)}} \\
 
\textbf{LoRA-FP Answer}: Janet's ducks lay 16 eggs per day.
She eats three for breakfast every morning, so she has 16 - 3 = 13 eggs left.
She bakes muffins for her friends every day with four eggs, so she has 13 - 4 = 9 eggs left. She sells the remainder at the farmers' market daily for \$2 per fresh duck egg, so she makes 9 * \$2 = \$18. Therefore, Janet makes \$18 every day at the farmers' market. The answer is \textcolor{blue}{18}. \\

\textbf{QLoRA Answer}: 
Janet's ducks lay 16 eggs per day.
She eats three for breakfast every morning, so she has 16 - 3 = 13 eggs left.
She bakes muffins for her friends every day with four eggs, so she has 13 - 4 = 9 eggs left. She sells the remainder at the farmers' market daily for \$2 per fresh duck egg, so she makes 9 * \$2 = \$18. Therefore, Janet makes \$18 every day at the farmers' market.
The answer is: \textcolor{blue}{18}  \\

\textbf{QA-LoRA Answer}: Janet's ducks lay 16 eggs per day.
She eats three for breakfast every morning, so she has 16 - 3 = 13 eggs left.
She bakes muffins for her friends every day with four, so she has 13 - 4 = 9 eggs left. She sells the remainder at the farmers' market daily for \$2 per fresh duck egg, so she makes 9 * \$2 = \$18. Therefore, Janet makes \$18 every day at the farmers' market. The answer is: \textcolor{blue}{18} \\

\textbf{QA-LoRA Answer}: Janet lays 16 eggs per day.
She eats three for breakfast, so she has 16 - 3 = 13 eggs left. She bakes muffins with four eggs, so she has 13 - 4 = 9 eggs left.  She sells the remainder at the farmers' market for \$2 per egg, so she makes 9 * \$2 = \$18.
Therefore, Janet makes \$18 every day at the farmers' market. The answer is: \textcolor{blue}{18} \\

\textbf{Our Answer}: Janet lays 16 eggs per day. She eats three for breakfast, so she has 16 - 3 = 13 eggs left. She bakes muffins with four eggs, so she has 13 - 4 = 9 eggs left. She sells the remainder at the farmers' market for \$2 per fresh duck egg, so she makes 9 * \$2 = \$18. Therefore, Janet makes \$18 every day at the farmers' market. The answer is: \textcolor{blue}{18}. 
\end{tcolorbox}

\clearpage
\begin{tcolorbox}[title={\large Case Study 2}, colback=cgray, colframe=cGreen, colbacktitle=cGreen, coltitle=white]
\textbf{Question}: 
A robe takes 2 bolts of blue fiber and half that much white fiber.  How many bolts in total does it take?
\textcolor{cGreen}{\textbf{(Ground-truth answer is 3)}} \\
 
\textbf{LoRA-FP Answer}: We know that the robe takes 2 bolts of blue fiber.
To find the amount of white fiber, we can use the ratio of blue fiber to white fiber, which is 2:1. So, the amount of white fiber is half of the blue fiber, which is 2/2 = 1 bolt. Therefore, the total number of bolts needed is 2 + 1 = 3 bolts. The answer is \textcolor{blue}{3}. \\

\textbf{QLoRA Answer}: 
The robe takes 2 bolts of blue fiber. 
To find the amount of white fiber, we can use the ratio of 1:2. 
Since the ratio is 1:2, we can set up the equation 1/2 * 2 = x, where x represents the amount of white fiber. 
Solving for x, we get x = 1. 
Therefore, the robe takes 2 bolts of blue fiber and 1 bolt of white fiber. 
In total, the robe takes 2 + 1 = 3 bolts of fiber. 
The answer is: \textcolor{blue}{3}  \\

\textbf{QA-LoRA Answer}:The robe takes 2 bolts of blue fiber. 
The robe takes half that much white fiber, which is 2/2 = 1 bolt of white fiber. 
So, in total, the robe takes 2 bolts of blue fiber + 1 bolt of white fiber = 3 bolts of fiber. 
Therefore, the robe takes 3 bolts of fiber in total.  The answer is: \textcolor{blue}{3} \\

\textbf{IR-QLoRA Answer}:
The robe takes 2 bolts of blue fiber. 
The robe takes half that much white fiber, which is 2/2 = 1 bolt of white fiber. 
So, the robe takes 2 bolts of blue fiber + 1 bolt of white fiber = 3 bolts in total. 
The answer is: \textcolor{blue}{3} \\

\textbf{Our Answer}: If a robe takes 2 bolts of blue fiber, then it takes 2/2 = 1 bolt of white fiber. So, the total number of bolts needed is 2 bolts of blue fiber + 1 bolt of white fiber = 3 bolts in total. The answer is: \textcolor{blue}{3}. 
\label{box:suppl-case-study}
\end{tcolorbox}

\end{document}